\documentclass[sn-mathphys,Numbered]{sn-jnl}

\usepackage{graphicx}%

\usepackage{amsmath,amssymb,amsfonts}%
\usepackage{amsthm}%
\usepackage{mathrsfs}%
\usepackage[title]{appendix}%
\usepackage{xcolor}%
\usepackage{textcomp}%
\usepackage{manyfoot}%
\usepackage{booktabs}%
\usepackage{algorithm}%
\usepackage{algorithmicx}%
\usepackage{algpseudocode}%
\usepackage{listings}%
\usepackage{subfigure}
\usepackage{natbib}
\usepackage{epsfig} 
\usepackage{mathptmx} 
\usepackage{times} 
\usepackage{amsmath} 
\usepackage{mathtools}
\usepackage{commath}
\usepackage{color}
\usepackage{threeparttable}
\usepackage{stfloats}
\usepackage{caption}
\usepackage{url}
\usepackage[switch]{lineno}
\usepackage{array}
\usepackage{makecell}        
\usepackage{multirow}%
\usepackage{tabularx}

\begin{document}

\title[SkipcrossNets: Adaptive Skip-cross Fusion for Road Detection]{SkipcrossNets: Adaptive Skip-cross Fusion for Road Detection}


\author[1,2,3]{\fnm{Yan} \sur{Gong}}\email{gongyan2020@foxmail.com}

\author*[1,]{\fnm{Xinyu} \sur{Zhang}}\email{xyzhang@tsinghua.edu.cn}

\author[2]{\fnm{Hao} \sur{Liu}}\email{liuhao163@jd.com}

\author[3]{\fnm{Xinming} \sur{Jiang}}\email{xinminj2001@gmail.com}

\author[1]{\fnm{Zhiwei} \sur{Li}}\email{2022500066@buct.edu.cn}

\author[1]{\fnm{Xin} \sur{Gao}}\email{skgxin@gmail.com}

\author[4]{\fnm{Lei} \sur{Lin}}\email{linlei@stu.xmu.edu.cn}

\author[1]{\fnm{Dafeng} \sur{Jin}}\email{jindf@tsinghua.edu.cn}

\author[1]{\fnm{Jun} \sur{Li}}\email{lijun19580326@126.com}

\author[1]{\fnm{Huaping} \sur{Liu}}\email{hpliu@tsinghua.edu.cn}

\affil[1]{\orgdiv{The State Key Laboratory of Automotive Safety and Energy, the School of Vehicle and Mobility}, \orgname{Tsinghua University}, \city{Beijing}, \state{100084}, \country{China}}

\affil[2]{\orgdiv{Autonomous Driving Division of X Research Department}, \orgname{JD Logistics}, \city{Beijing}, \postcode{101111}, \country{China}}

\affil[3]{\orgname{School of Computer Science and Engineering, Northeastern University}, \city{Shenyang}, \postcode{110169}, \country{China}}

\affil[4]{\orgname{Department of Artificial Intelligence, School of Informatics, Xiamen University}, \city{Xiamen}, \postcode{361005}, \country{China}}


\abstract{Multi-modal fusion is increasingly being used for autonomous driving tasks, as different modalities provide unique information for feature extraction. 
However, the existing two-stream networks are only fused at a specific network layer, which requires a lot of manual attempts to set up. As the CNN goes deeper, the two modal features become more and more advanced and abstract, and the fusion occurs at the feature level with a large gap, which can easily hurt the performance.
To reduce the loss of height and depth information during the process of projecting point clouds into 2D space, we utilize calibration parameters to project the point cloud into Altitude Difference Images (ADIs), which exhibit more distinct road features.
In this study, we propose a novel fusion architecture called Skip-cross Networks (SkipcrossNets), which combine adaptively ADIs and camera images without being bound to a certain fusion epoch. 
Specifically, skip-cross fusion strategy connects each layer to each layer in a feed-forward manner, and for each layer, the feature maps of all previous layers are used as input and its own feature maps are used as input to all subsequent layers for the other modality, enhancing feature propagation and multi-modal features fusion.
This strategy facilitates selection of the most similar feature layers from two modalities, enhancing feature reuse and providing complementary effects for sparse point cloud features.
The advantages of skip-cross fusion strategy is demonstrated through application to the KITTI and A2D2 datasets, achieving a MaxF score of 96.85\% on KITTI and an F1 score of 84.84\% on A2D2. The model parameters require only 2.33 MB of memory at a speed of 68.24 FPS, which can be viable for mobile terminals and embedded devices.}





\maketitle

\section{Introduction}\label{sec1}

Road detection is not only an important prerequisite for vehicle trajectory planning and decision-making but also a key technology in autonomous driving. Recent studies have generally used pixel-level semantic segmentation for road detection, which can provide effective road information at the pixel level \cite{kang2011multiband, zhang2021multi, gong2024tclanenet, wan2024adnet}.

The camera captures light in the environment to provide high-resolution images, enabling the recognition of object appearances, shapes, colors, and textures.
Many camera-based methods have made significant progress and achieved good results \cite{bandara2022spin, chang2022fast}. However, cameras perform poorly in low-light environments and are difficult to use for distance and depth perception \cite{alvarez2012road, zhou2024decoupling, wan2023lfrnet}. 
In contrast, LiDAR is based on the reflection of surrounding obstacles and provides complementary 3D point cloud information that is not affected by ambient light. Although LiDAR point cloud data are typically sparse, they can be used to accurately measure distances and are less affected by lighting \cite{PLARD}. 
However, the cost of LiDAR is high, and in bad weather such as rain and snow, the performance of LiDAR will decrease, reducing its application range \cite{tomas2016forecasting, zhang2023multi, gong2023sifdrivenet}. 
In conclusion, both LiDAR and camera have their unique strengths and limitations, and considering the fusion of these two modalities can provide more complementary information and improve the overall reliability of road detection.
	
	\begin{figure}[]
	    \centering
	    
	    \subfigure[A comparison of model parameters and performance on A2D2]{
		\begin{minipage}[t]{\linewidth}
			\centering
			\includegraphics[width=0.9\linewidth]{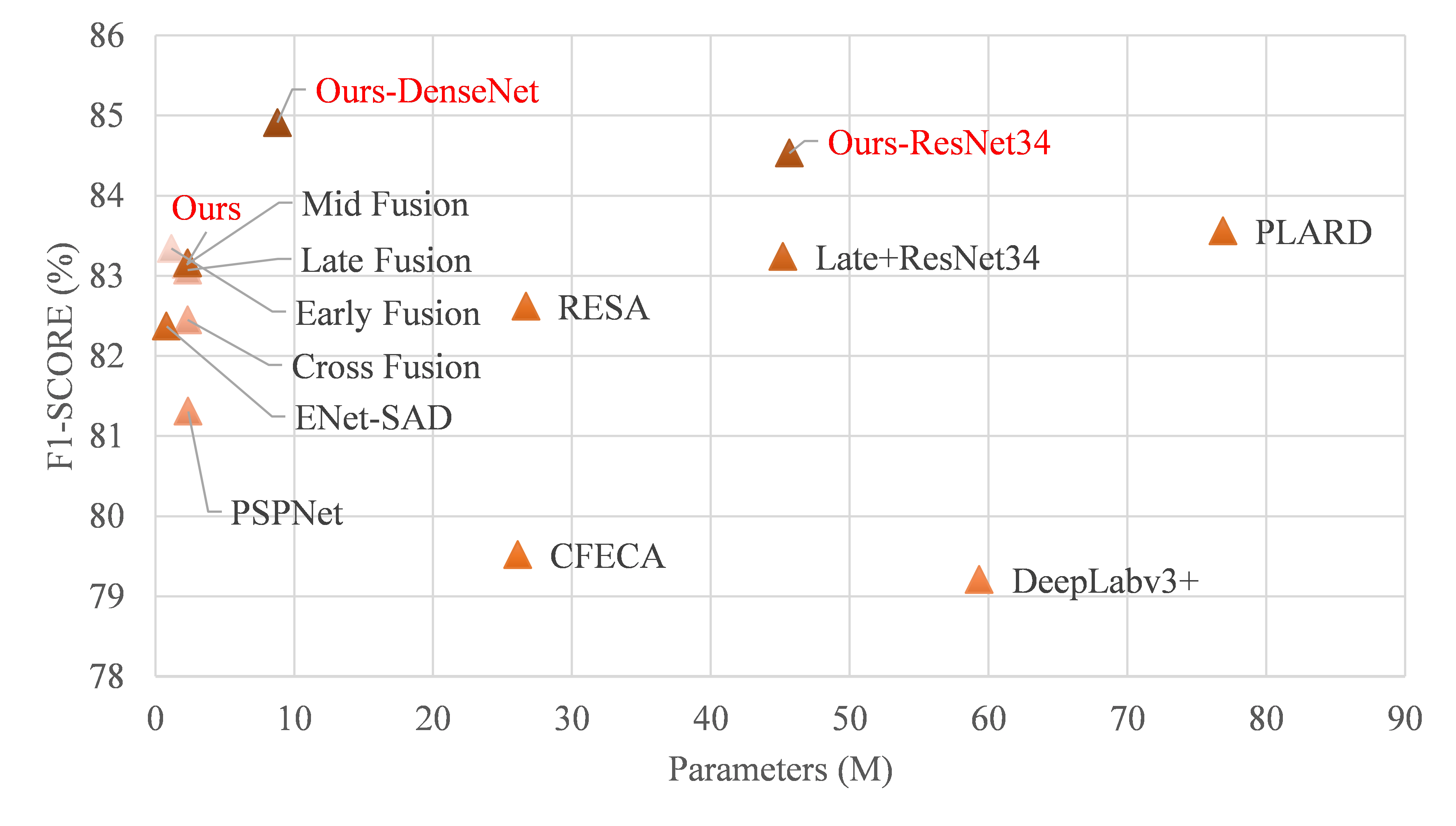}
			\end{minipage}%
		}%

		\subfigure[A comparison of model parameters and performance on KITTI]{
		\begin{minipage}[t]{\linewidth}
			\centering
			\includegraphics[width=0.9\linewidth]{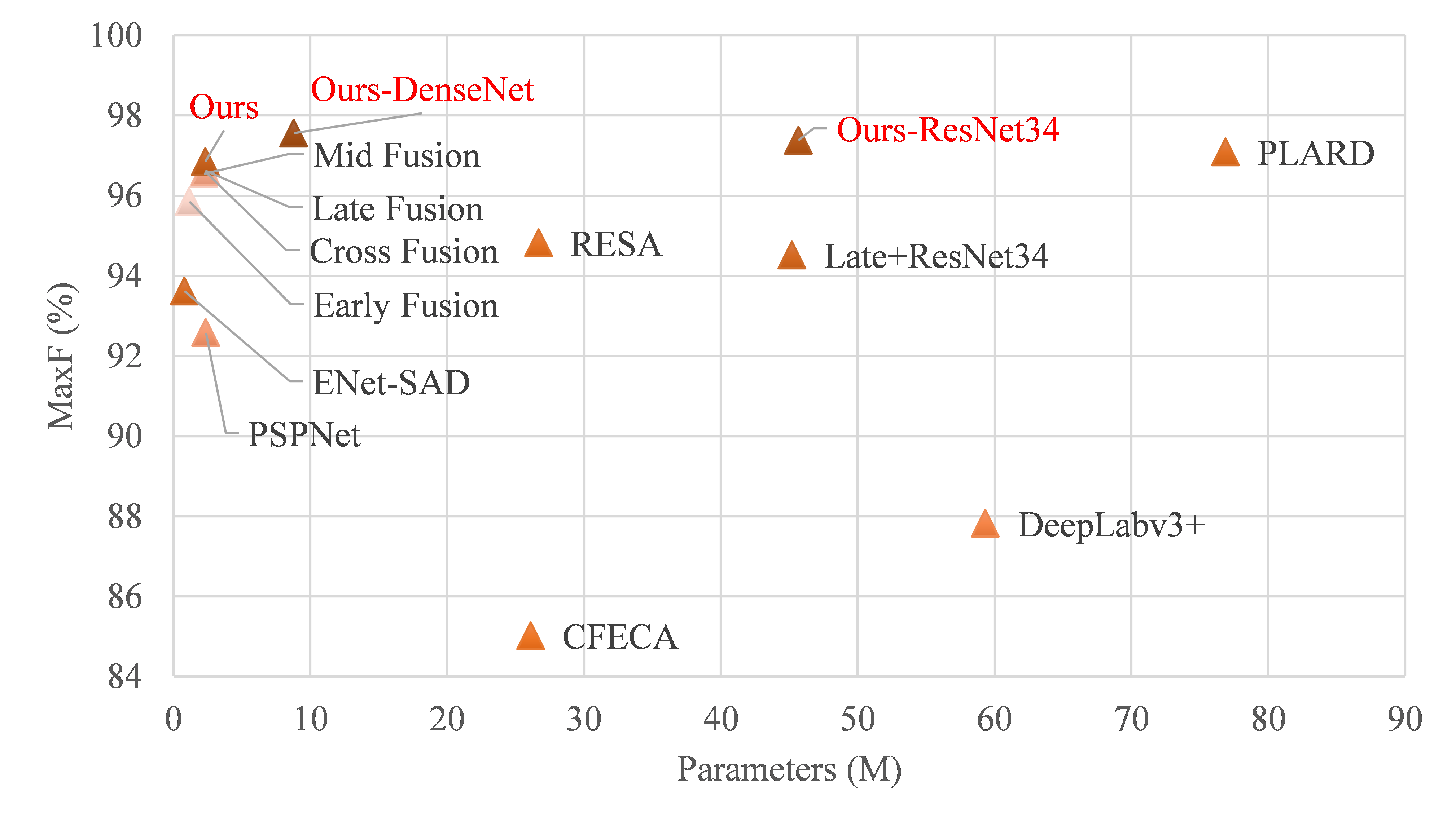}
			\end{minipage}%
		}%
	    \caption{We compare the model parameters and performance of various methods on KITTI and A2D2 scenarios. The red mark part is our proposed method.}
	    \label{fig:fpsmaxf}
	\end{figure}

	Unfortunately, fusing LiDAR and camera images is a challenge for multi-modal fusion \cite{feng2020deep} due to the differences in data space and feature space for different modalities.
    For the data space, the difficulty of integrating these two data types lies in the fact that LiDAR operates in a 3D space while cameras operate in a 2D space, making it challenging to define a suitable space to integrate these two types of data. Inconsistent data spaces can lead to poor performance of the models \cite{PLARD}.
    For feature space, most of the existing multi-modal methods focus on the process of ``how to fuse'', but lack a thorough explanation for the period of ``when to fuse''.
    The fusion period refers to the period when the two-stream network executes the multi-modal feature fusion step, which has a significant impact on algorithm performance, as shown in Fig.~\ref{fig:Fusion_Stages}. Existing strategies have been used to fuse source data, feature maps or outputs at different fusion periods, corresponding to early, middle and late fusion, respectively \cite{gong2023feature}. Early fusion \cite{wulff2018early} requires a high degree of spatiotemporal alignment in the original data. In middle fusion \cite{xu2019mid}, it is difficult to determine which feature layer in the backbone should be selected to perform the fusion step. Late fusion \cite{schlosser2016fusing, liu2024glmdrivenet} focuses on the results of two-stream network fusion at the decision layer and may ignore intermediate semantic features.
    As neural networks get deeper and features become more abstract and high-level, the feasibility of merging this data depends heavily on whether the features in both sources exist at the same level. 
    However, the fusion periods of existing multi-modal fusion methods are manually set, making it challenging to determine the optimal timing for different modalities fusion.

    \begin{figure*}[!htb]
		\centering
		\includegraphics[width=1 \linewidth]{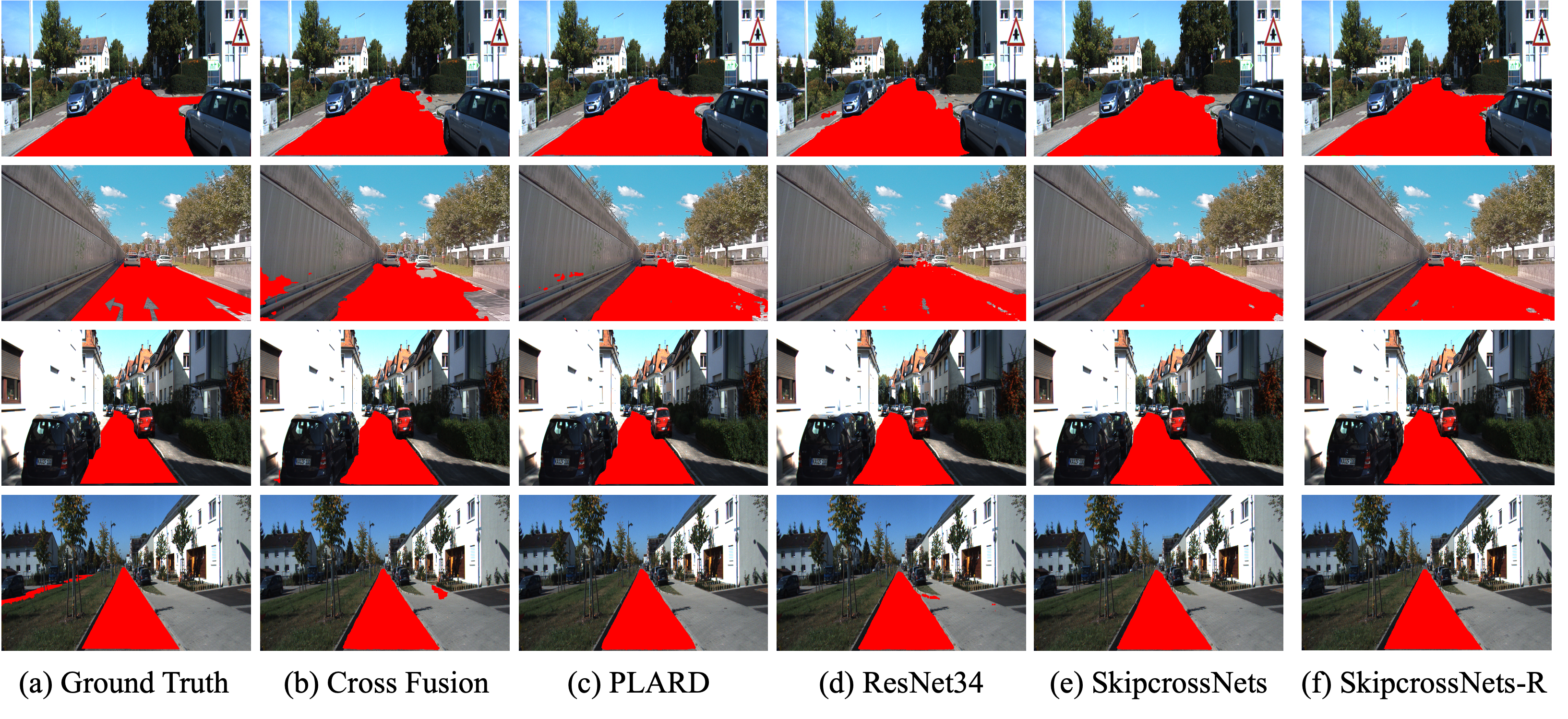}
		\caption{Results of State-of-the-art models and ours: the first two rows are tested on the KITTI dataset and the rest are tested on the A2D2 dataset. For the sub-pictures (b) and (d), it is obvious that the resulting picture is quite different from the Ground Truth. Sub-pictures (c) have more noise in the second picture. Overall, the third row has a better performance, but for the edges of objects, such as the cars in the third row, the results of SkipcrossNets are smoother.}
		\label{fig:introduction_vis}
	\end{figure*}
	
	In this paper, we propose a novel fusion strategy, skip-cross fusion, for point clouds and visual images. Skip-cross fusion connects the feature maps of each modality to all feature maps after the corresponding position of the other modality, realizing a dense cross-modal connection. In this process, skip-cross blocks are used to adaptively identify the best period combinations of LiDAR point clouds and visual images. SkipcrossNets then make full use of multi-modal information in the encoder stage and thereby achieve superior performance for sparse point clouds. 
	At the same time, we also use skip-cross fusion in the decoder, so that each feature map uses another modal feature map in the process of restoring the resolution.
	For data space, we reduce the gap between LiDAR and visual data spaces by projecting LiDAR as an altitude difference image (ADI), since the roads are correlated with height information in urban scenes. 
	
	Fig.~\ref{fig:fpsmaxf} provides a comparison of various techniques, evaluated in terms of performance and model parameter quantities. SkipcrossNets-R and SkipcrossNets-D represent models utilizing ResNet34 \cite{resnet} and DenseNet \cite{huang2019convolutional} as the backbone structure, respectively. Early \cite{wulff2018early}, middle \cite{xu2019mid}, and late fusion \cite{schlosser2016fusing} are described in Section \ref{sec3}, along with cross-fusion \cite{caltagirone2019lidar}. Early fusion is the lightest model, but its performance is inferior to many existing methods. PSPNet \cite{PSPNet} and DeepLabv3+ \cite{DeepLabv3+} include MobileNet and Resnet101 as backbones. 
    Performance comparisons suggest SkipcrossNets-R and the Progressive LiDAR Adaptation for Road Detection (PLARD) \cite{PLARD} are sufficiently competitive for the KITTI \cite{geiger2012we} dataset, with PLARD ranking highest. However, the number of parameters and calculations required by SkipcrossNets-R are only 59\% and 57\% of those in PLARD. Thus, when considering performance and complexity to be of equal importance, SkipcrossNets outperform other models. In addition, the SkipcrossNets series models achieve the best performance for the A2D2 dataset. SkipcrossNets also adopt a dense structure between the two modalities similar to DenseNets \cite{huang2019convolutional}, which significantly reduce the number of parameters (2.33 MB). This in turn improves the information flow and gradients in the network, which simplifies training. In this study, SkipcrossNets are compared with other fusion strategies and state-of-the-art (SOTA) methods applied to two benchmark datasets: KITTI \cite{geiger2012we} and A2D2 \cite{baltruvsaitis2018multimodal}.
As shown in Fig. \ref{fig:introduction_vis}, the results in columns (b), (c), and (d) exhibit certain discrepancies from their corresponding ground truths. The results of SkipcrossNets and SkipcrossNets-R demonstrate better performance with results that are smoother along object edges (e.g., cars), which demonstrate the benefits of our skip-cross fusion strategy. Overall, SkipcrossNets significantly outperform current state-of-the-art algorithms for most benchmark tasks and has significant potential for autonomous driving applications.
	
	The contributions of this study can be described as follows:
    
	 \begin{itemize}
		\item A novel skip-cross fusion strategy is proposed to perform feature fusion at each layer of the two-stream's network and adaptively select the best fusion period, instead of only occurring in a specific epoch as before. At the same time, we also found that early, middle, and late fusion are just a special form of skip-cross fusion.
		\item Since the heights of roads, vehicles, and buildings in 3D space are different, we project LiDAR to the image plane to generate altitude difference images to better distinguish road areas and reduce fusion problems caused by spatial differences.
		\item To tackle the issue of large parameter size in existing multi-modal fusion methods, we develop the lightweight SkipcrossNets with the small parameter size of 2.33 MB and a fast speed of 68.24 FPS. In addition, our proposed method performs well in sparse point cloud scenes, which can mitigate the issue of costly high-line LiDAR and adequately meet the needs of autonomous driving.
	\end{itemize}
	
	In the following sections, we first review recent progress in road detection and multi-modal fusion. We then demonstrate the proposed skip-cross fusion network architecture and compare it with other fusion strategies applied to the KITTI and A2D2 road datasets. Finally, the results of a quantitative analysis and an ablation study are presented and discussed.

\section{Related Work}\label{sec2}

The related work of skip-cross fusion is introduced in the following sections. 
Section \ref{trd} describes the conventional road detection. 
The visual image-based road detection is introduced in Section \ref{bird}.
Section \ref{mmf} describes the latest methods of multi-modal fusion with LiDAR and camera.

\subsection{Traditional Road Detection}
\label{trd}
Existing lane detection algorithms typically employ geometric information and image processing (e.g., HSI color models \cite{sun2006hsi} and edge extraction \cite{yu1997lane, wang2000lane}) to enhance road connectivity by using visual cues and combining context from prior conditions including road geometry \cite{laptev2000automatic}, high-order CRF formulas \cite{wegner2013higher}, marking point processes \cite{chai2013recovering, stoica2004gibbs}, and planning problem solutions. Qi Wang et al. \cite{wang2015adaptive} proposed a context-aware road detection model that combined depth cues with a temporal MRF model, utilizing traditional RGB colors and label transfer in an efficient nearest neighbor search process that could distinguish road and non-road areas. In addition, Markov and conditional random fields were also used in post-processing steps. However, these methods require manual design of features and artificially set thresholds, are highly empirical, and are less effective for road detection in complex scenes.

\subsection{Visual Image-based Road Detection}
\label{bird}
High-precision road detection has been made possible by continued advances in computing power, allowing for its deployment in the field of autonomous driving \cite{qi2018dynamic, chen2018parallel, kong2009vanishing, zhang2022openmpd, chen2017generic, zhang2023oblique}. Prior road detection research has largely been conducted using pixel-wise semantic segmentation based on deep learning. For example, Teichmann et al. \cite{teichmann2018multinet} adopted a convolutional neural network (CNN) \cite{he2016accurate} for road detection with monocular camera images. However, this approach is easily affected by dynamic illumination conditions. Long et al. \cite{long2015fully} proposed fully convolutional and upsampling layers to solve pixel-level semantic segmentation problems. Lee et al. \cite{lee2017vpgnet} proposed a VPGNet to study detection in a variety of weak lighting conditions and shaded regions considering contextual information in the data. Zou et al. \cite{zou2019robust} proposed ConvLSTM to improve detection performance in specific cases, e.g., vehicle occlusion and tree branch shadows. Alvarez et al. \cite{alvarez2014combining} used contextual cues for road detection, including horizon lines, vanishing points, lane markings, and road geometry, all of which are robust to varying conditions. Yang et al. \cite{yang2021capturing} captured the inherent remote dependencies in omnidirectional data using cross  $360^{o}$ images and prior contextual knowledge (based on an attention mechanism) to improve model training. Road detection has also been approached as a special case of panoramic semantic segmentation. For instance, Yang et al. \cite{yang2019pass} proposed a panoramic annular semantic segmentation model to perceive whole environments, solving boundary discontinuities and overlapping semantic problems. Yang et al. \cite{yang2020omnisupervised} proposed an omni-supervised learning framework based on efficient CNNs, which achieved significant universality gains in panoramic images. But single-model methods are easily affected by external conditions and have poor stability. Although the camera provides high-resolution images, it fails in bad scenes such as low light, night, and exposure. The limitations brought by the sensor itself are difficult to solve through the design of the method, so consider introducing other modal data to assist.

\subsection{Multi-modal Fusion}
\label{mmf}
Conventional road segmentation algorithms use LiDAR and camera fusion to improve segmentation effects and compensate for the inherent shortcomings of single data sources. The primary technical problem of then involves achieving effective multi-sensor fusion in real-time. Most techniques involve data preprocessing and the projection of LiDAR to a 2D format, such that both the camera and LiDAR are consistent in the data space(i.e., the BEV, height map, and contour map).
Xiao et al. \cite{xiao2018hybrid} proposed a novel hybrid conditional random field (CRF) model that accepts the labels (road or background) of aligned pixels and LIDAR points as random variables, inferring labels by minimizing a hybrid energy function.
Caltagirone et al. \cite{caltagirone2017fast} proposed a fully convolutional network for road detection with 2D top-view images transformed from LiDAR point clouds used as input.
Wang et al. \cite{wang2017embedding} proposed a Siamese fully revolutionary network to process RGB images and semantic contours utilizing the prior knowledge from location lines to improve the final lane detection effect.
Caltagirone et al. \cite{caltagirone2019lidar} proposed a network, which outperformed comparable fusion strategies, in which cross fusion occurs in the same layer level while features in different levels may not be fully used for road detection.
Chen et al. \cite{PLARD} used a transfer function to adapt a LiDAR feature space to a visual feature space to better supplement and improve features. Visual information was then integrated with the adjusted LiDAR information. 
Zhang et al. \cite{CFECA} proposed a channel attention mechanism used to obtain local interaction information across channels. Weights were assigned to represent the contributions of different feature channels for the LiDAR and camera.
Samal et al. \cite{SamalKSWM22} proposed a spatiotemporal sampling algorithm that activates Lidar only at regions of interest identified by analyzing visual input. The algorithm significantly reduces Lidar usage.
A hierarchical multimodal fusion module is developed by Zhou et al. \cite{ZhouDLY23} to enhance feature fusion, and a high-level semantic module is constructed to extract semantic information, so as to fuse with rough features at various levels of abstraction. RangeSeg \cite{ChenC22} used a shared encoder backbone with two range-dependent decoders. The heavy one only calculates the distance and the top of the distance image where the small object is located to improve the detection accuracy of the small object, while the light one only calculates the entire distance image to reduce the calculation cost. 
By combining case segmentation with lane center estimation, Sun et al. \cite{SunLXS23} proposed an adaptive multi-lane detection method for intelligent vehicles based on instance segmentation, which incorporates cosine measurement into the loss function, so that the proposed method can extract more discriminative features for foreground extraction.

Most of these fusion methods are fused in a specific period, and different periods are artificially set, which is very empirical. Therefore, we propose an adaptive skip-cross fusion strategy embedded between each layer of the backbone to avoid the above-mentioned problems with the fusion period, which significantly increases the proportion of fused information and improves the accuracy and robustness of road detection.
 
	


\begin{figure}[]
    \centering
    \includegraphics[width=0.8\linewidth]{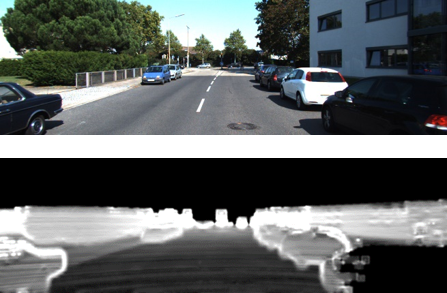}
    \caption{The image pairs of RGB and Altitude Difference Image (ADI).}
    \label{fig:RGB_and_ADI}
\end{figure}

\section{Methodology}\label{sec3}

The difficulty in the fusion of LiDAR and visual images lies in the differences between their data spaces and the selection of appropriate fusion periods and techniques.
In Section \ref{Dataspace}, we describe the data spaces for both LiDAR and visual images. In Section \ref{Network_architecture}, we report on the basic network architecture. In Section  \ref{SK}, we compare other fusion strategies and propose skip-cross fusion.

\subsection{Data Spaces}
\label{Dataspace}
	
	LiDAR point clouds and visual images span different dimensions, and a common fusion approach involves projecting point cloud data onto a 2D plane for a simpler combination with visual image pixels. However, this projection process often leads to losses in height and depth information, especially in fine-grained tasks such as lane detection. 
	
	To make road features more obvious, we generate single-channel altitude difference images (ADIs), which are converted from LiDAR point clouds.
	As shown in Fig.~\ref{fig:RGB_and_ADI}, LiDAR data in the form of 3D coordinate vectors can be projected onto a 2D image plane using the provided calibration parameters. The intensities of 2D image pixels then represent normalized X, Y, and Z coordinates. Altitude differences in the spatial offsets between two positions (${Z_{x,y}}$ and $N_{x},N_{y}$), are then calculated from the LiDAR data to obtain ADIs for road features. This process can be represented as:
	\begin{equation}	
	V_{x,y}=\frac{1}{M}\sum_{N_{x},N_{y}}\frac{\left | Z_{x,y}-Z_{N_{x},N_{y}} \right |}{\sqrt{(N_{x}-x)^{2}+(N_{y}-y)^{2}}} ,
	\end{equation}
	where ${Z_{x,y}}$ is the altitude of the LiDAR point projected onto ($x$, $y$), ($N_{x},N_{y}$) is the position of ($x$, $y$) in the neighborhood, $V_{x,y}$ is the projected pixel value at ($x$, $y$), and $M$ is the total number of considered neighborhood positions \cite{PLARD}. The background is darkest in ADIs, while obstacles are the brightest, and lane areas are generally located in the darker regions in the lower half of the image. The pixel values represent object altitudes relative to the road plane and can be used to distinguish road areas from non-road areas. 
	
	\begin{figure*}[tp]
		\centering
		\includegraphics[width=1\linewidth]{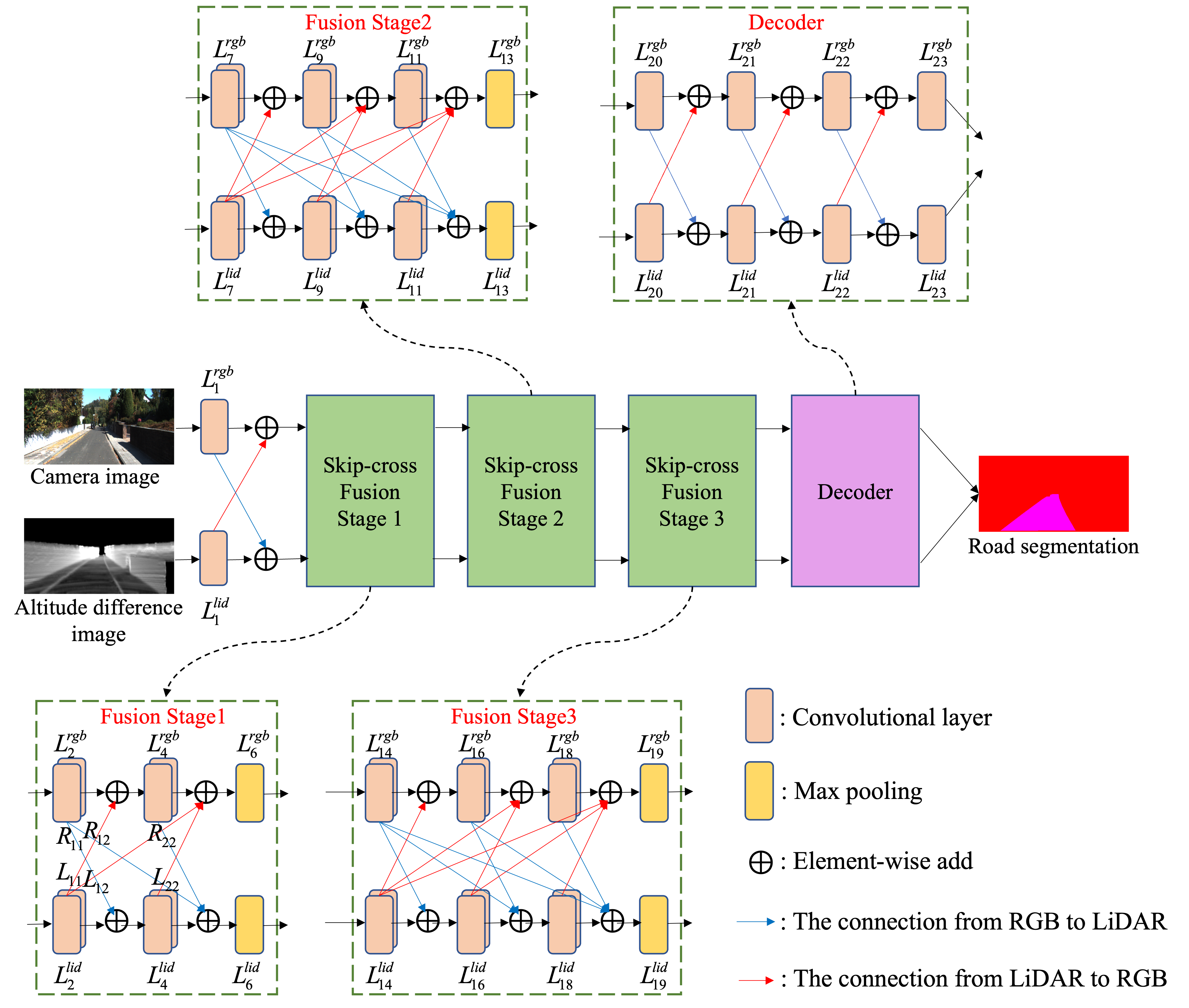}
		\caption{The overall architecture of the SkipcrossNets. The encoder includes three fusion stages, and each fusion stage adopts the skip-cross fusion strategy detailed in Section \ref{SK}. Skip-cross fusion directly integrates information from the feature extraction and processes it by training cross-connected branches, which can be fused at any depth, not just limited to a certain level compared with other fusion strategies as shown in Fig. \ref{fig:Fusion_Stages}.}
		\label{fig:Single_Modal}
	\end{figure*}
	
	\subsection{Network Architecture}
	\label{Network_architecture}
	The basic network used in this study are comprised of a fully convolutional encoder-decoder with $3\times3$ kernels as shown in Fig.~\ref{fig:Single_Modal}. The network consists of 23 layers, including encoder L$_2$-L$_{19}$ and decoder L$_{20}$-L$_{23}$. The encoder consists of a single convolution layer with a stride of 2 and three fusion stages, and each fusion stage includes several BasicBlocks of ResNet \cite{resnet} and ended with a $2\times2$ max pooling layer. The encoding module downsamples four times, and the decoding module includes four transposed convolution layers that conduct upsampling and restore the feature map to its original input size. The included transposed convolution layers are set to a 3$\times$3 kernel size and a stride of 2. Each convolution layer is then followed by a rectified linear unit (ReLU) layer. The activation function performs gradient descent and backpropagation, avoiding the problem of gradient explosion and disappearance. In addition, skip-connections are used in the encoder and decoder steps to recover detailed features and restore resolution.


	\begin{figure}[htbp]
		\centering
		\subfigure[Early fusion]{
			\begin{minipage}[t]{1\linewidth}
				\flushleft 
				\includegraphics[width=2.8in]{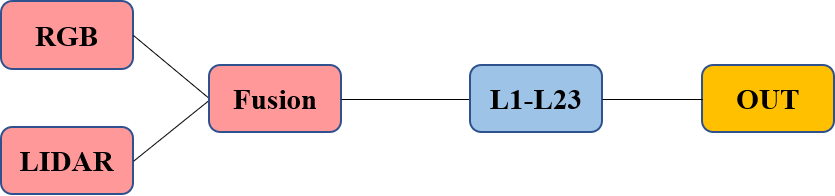}
                \centering
			\end{minipage}%
		}%
		
		\subfigure[Middle fusion]{
			\begin{minipage}[t]{1\linewidth}
				\flushleft 
				\includegraphics[width=3.65in]{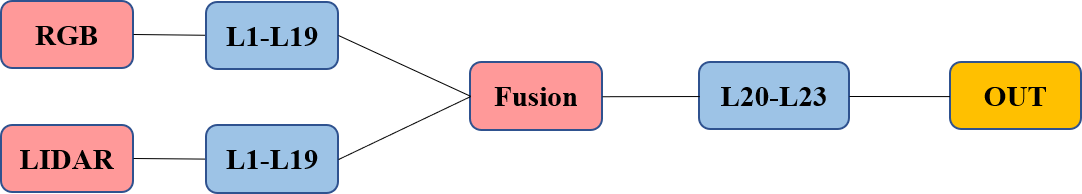}
                \centering
			\end{minipage}%
		}%
		
		\subfigure[Late fusion]{
			\begin{minipage}[t]{1\linewidth}
				\flushleft 
				\includegraphics[width=2.9in]{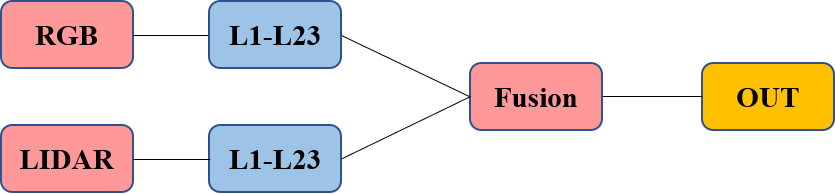}
                \centering
			\end{minipage}
		}%
		
		
		\caption{(a) Early fusion (b) Middle fusion (c) Late fusion. The main differences of the three fusion strategies are in the fusion stages. When a fusion occurs in a deeper network, the features participating in the fusion become more abstract and the model is more flexible. However, effective intermediate and detailed features may also be lost.}

		\label{fig:Fusion_Stages}
	\end{figure}
	
	\subsection{Skip-cross Fusion}
	\label{SK}
	The varying degrees of feature abstraction caused by uncertainty in multi-modal fusion time make it difficult to achieve robust road detection using only a specific fusion type for LiDAR point clouds and visual images, as shown in Fig. \ref{fig:Fusion_Stages}. 
	To overcome this problem and improve road detection, we propose an adaptive skip-cross fusion that effectively integrates LiDAR data and visual information.
	
	The overall design of SkipcrossNets follows a common encoder-decoder structure. The encoder stage is more important for the multi-modal information fusion process. In the encoder stage, sufficient fusion must be ensured to generate a better intermediate feature map and restore the resolution and detailed features in the image. The encoder includes three fusion stages each of which stage contains a specified number of residual blocks (e.g., 2, 3, and 3 in the subsequent experimental setup). A basic ResNet structure are adopted, namely, two 3$\times$3 convolution layers and one identity shortcut. The first fusion stage is composed of two residual blocks and one max pooling layer while the other two fusion stages are composed of three residual blocks and one max pooling layer. 
	The basic structure of SkipcrossNets in Fig.~\ref{fig:Single_Modal} includes three stages, each stage includes multiple densely connected blocks.
    To avoid the problems caused by different fusion periods, we perform cross-modal dense connections in the two-stream network and pass the feature map of each modality to all feature maps after the corresponding position of the other modality through a learnable factor. Different from the previous fusion only at a certain network layer, the skip-cross fusion strategy increases the feature interaction of different modals. The specific layers where multi-modal fusion occurs depend on the learning of cross-modal learnable factors, rather than artificial settings, such fusion networks have better performance. Therefore, a traditional two-stream convolutional network with $2L$ layers has $2L$ connections, one connection between each layer and its subsequent layers, while our network has $(L+1)L$ direct connections, occurring in its all subsequent layers.
	As we consider a more complex learning model, denser connections are adopted in each stage and each layer is feedforward connected to other layers. In the decoder structure, skip-fusion is used to perform weighted cross-connection to the fused feature maps to improve upsampling and restore the original resolution. The proposed fusion model can also provide more effective and in-depth training, thereby strengthening feature transfers and improving feature reuse rates. 

	The first fusion stage, the detailed architecture as shown in Fig.~\ref{fig:Single_Modal}, includes LiDAR and RGB branches. The two LiDAR and RGB blocks are denoted $\mathbf{L}^{lid}_{k}$ and $\mathbf{L}^{rgb}_{k}$  respectively. In the following expressions, L$_{kj}$ presents the weight from the $k$-th LiDAR block to the $j$-th RGB block and R$_{kj}$ denotes the weight from the $k$-th RGB block to the $j$-th LiDAR block. Other stages are similar to the first fusion stage, and the first fusion stage is detailed as follows:
    
	\begin{equation}
	\begin{split}
    \mathbf{L}^{lid}_{4} &  = \mathbf{L}^{lid}_{2} + \mathrm{R11} \times \mathbf{L}^{rgb}_{2}, \\    \mathbf{L}^{rgb}_{4} &  = \mathbf{L}^{rgb}_{2} + \mathrm{L11} \times \mathbf{L}^{lid}_{2},    \\ 
	 \mathbf{L}^{lid}_{6} &  = \mathbf{L}^{lid}_{4} + \mathrm{R12} \times \mathbf{L}^{rgb}_{2} + \mathrm{R22} \times \mathbf{L}^{rgb}_{4}, \\ 
	 \mathbf{I}^{rgb}_{6} & = \mathbf{L}^{rgb}_{4} + \mathrm{L12} \times \mathbf{L}^{lid}_{2} + \mathrm{L22} \times \mathbf{L}^{lid}_{4}.
	\end{split}
	\end{equation}

	Early, middle, and late fusion (see Fig. \ref{fig:Fusion_Stages}) are special cases of skip-cross fusion. The basic principle behind this strategy is to allow a fully convolutional network (FCN) to process any depth information, rather than limiting it to a certain level as mentioned previously. In the training process, these fusion parameters are learned using a loss function to better integrate LiDAR and visual images. 
	The high-level features include rich semantic information, which can improve classification accuracy. The low-level features include detailed information which can increase positioning accuracy. Skip-cross fusion fully considers feature fusion between different levels, thus increasing information flow and fully reusing these features. It also transforms the selection of fusion methods into the learning of network parameters, which enables the network to learn fusion methods both flexibly and autonomously. By introducing a dense structure, SkipcrossNets are compatible with more complex fusion models, thereby reducing the number of parameters, and making the network easier to train.

\section{Experimental Results}
	
	The effectiveness of SkipcrossNets is evaluated through a sequence of validation experiments as outlined in the following sections. Section \ref{dataset} describes the characteristics and preprocessing steps on KITTI and A2D2 datasets. The detailed training process is recounted in Section \ref{IAT}, while Section  \ref{KITTI_e} and \ref{A2D2_e} present a comparison of SkipcrossNets with other state-of-the-art methods, applied on KITTI and A2D2 datasets respectively. Section \ref{CK} provides the results of cross-dataset knowledge experiments, and finally, Section \ref{AS} offers the results of ablation studies.

	\subsection{Dataset}
    \label{dataset}
	\subsubsection{Overview}
	Multiple algorithms are applied to the KITTI and A2D2 datasets. The fusion networks are trained using KITTI road data \cite{xu2019mid}, consisting of 289 training images and 290 test images acquired over several days in city, rural, and highway scenes. Specifically, these images include three scene categories: marked urban roads (191 samples), multiple marked urban roads (190 samples), and unmarked urban roads (198 samples). Real labels in the test images are not opened, and the results can only be submitted once to evaluate algorithm performance with the test set. As such, we divide the images into training and validation sets using specific ratios. The performance of several different algorithms is evaluated using this validation set. Since images in the datasets do not exhibit a uniform resolution, resizing operations are applied to ensure each image are consistently 1280$\times$384 pixels. The three-channel label images are also converted to single-channel binary images.


	A2D2 data are acquired by six cameras and five LiDAR units for a full 360-degree coverage. The recorded data is time-synchronized and registered with other data and annotations are added for non-continuous frames (e.g., 41,277 frames with semantic split images and point cloud labels). The non-annotated sensor data consists of 392,556 consecutive frames acquired in three cities in southern Germany. KITTI is comprised of 64-line LiDAR data used to generate point clouds, while A2D2 combines one 8-line and two 16-line LiDARs. Since A2D2 point cloud images are overly sparse, we randomly select a subregion for upsampling, which is then output as a 2D projection image. We also select the road and normal streets to establish road detection annotations. The number of pictures in the training, test, and verification sets is 470, 238, and 78, respectively.

	\subsubsection{Performance Evaluation}
	Evaluation metrics calculated for the KITTI benchmark include maximum F-measure (MaxF), average precision (AP), precision (PRE), recall (REC), false positive rate (FPR), and false negative rate (FNR) \cite{fritsch2013new}. F-measure derived from precision and recall values for pixel-based evaluations is calculated as:
	
	%
	%
	\begin{equation}
	\label{equ:F2}
	\emph{F-measure} =  \left(1+\beta^{2}\right) \frac{\emph{PRE} \times \emph{REC}}{\beta^{2}\emph{PRE + REC}},
	\end{equation}
	%
	%
	
	%
	where $\beta$ = 1 for F1, and $\beta$ = 2 for F2. Other evaluation indicators are also used in the A2D2 benchmark, including precision (PRE), recall (REC), F1, F2, mean intersection over union (MIOU), and accuracy (ACC). These metrics can be defined as:
	\begin{equation}
	\emph{MIOU} = \frac{1}{k + 1} \sum\limits_{i = 0}^{k} \frac{\emph{TP}}{\emph{FN + FP + TP}},
	\end{equation}
	
	\begin{equation}
	\emph{ACC} = \frac{\emph{TP + TN}}{\emph{TP + TN + FP + FN}},
	\end{equation}
	where $k$ + 1 represents the number of categories (including background), and TP, TN, FP, and FN denote true positive, true negative, false positive, and false negative respectively.

	\begin{table}[!htb]
		
		\caption{A performance comparison of fusion algorithms applied to three KITTI road scenes in the validation set. The ``$\uparrow$'' indicates a larger numerical value, reflecting better results, while the ``$\downarrow$'' signifies the opposite.}
		\label{tab:uu}
        \def\arraystretch{1.2}
        \item 
		\begin{tabular}{c|llllllll}
                \hline
                \multirow{2}{*}{Model} & \multicolumn{6}{c}{URBAN MARKED ROAD SCENE}        \\  
                & MaxF $\uparrow$  & AP  $\uparrow$  & PRE $\uparrow$ & REC $\uparrow$ & FPR $\downarrow$ & FNR $\downarrow$     \\ \hline
                
                Early	Fusion	&	95.24	&	87.08	&	94.17	&	96.32	&	1.15	&	3.68	\\	
                Mid	Fusion	&	96.54	&	88.81	&	96.08	&	97.00	&	0.76	&	3.00	\\	
                Late	Fusion	&	96.56	&	88.99	&	96.27	&	96.85	&	0.72	&	3.15	\\	
                Cross	Fusion	&	96.53	&	89.13	&	96.43	&	96.63	&	0.69	&	3.37	\\	
                Late+ResNet34	&	94.79		&	86.65		&	93.7		&	95.91		&	1.24		&	4.09 \\	
                
                PLARD    & 97.17 & \textbf{92.35} & 97.87 & 96.67          & \textbf{0.41} & 3.33        \\ 
                \hline
                PSPNet	&	90.43		&	83.44		&	90.17		&	90.69		&	1.91		&	9.31 \\
                DeepLabV3+	&	87.29		&	74.39		&	80.22		&	95.73		&	4.55		&	4.27 \\
                \hline
                SkipcrossNets	&	96.85	$\uparrow$	&	89.26	$\uparrow$	&	96.57	$\uparrow$	&	\textbf{97.14}	$\uparrow$	&	0.67	$\downarrow$	&	2.86	$\downarrow$	\\	
                SkipcrossNets-R	&	97.14	$\uparrow$	&	91.75	$\uparrow$	&	97.38	$\uparrow$	&	96.82		&	0.73	 	&	2.58	$\downarrow$	\\
                SkipcrossNets-D	&	\textbf{97.38}	$\uparrow$	&	91.66	$\uparrow$	&	\textbf{97.91}	$\uparrow$	&	96.66		&	0.80	 	&\textbf {2.34}	$\downarrow$	\\
                
                \hline \hline
                
                \multirow{2}{*}{Model} &  \multicolumn{6}{c}{URBAN MULTIPLE MARKED ROAD SCENE}      \\ 
                & MaxF $\uparrow$  & AP  $\uparrow$  & PRE $\uparrow$ & REC $\uparrow$ & FPR $\downarrow$ & FNR $\downarrow$ \\ \hline
                
                Early	Fusion	&	96.45	&	90.26	&	96.95	&	95.96	&	0.92	&	4.04	\\	
                Mid	Fusion	&	96.52	&	91.11	&	97.88	&	95.19	&	0.63	&	4.81	\\	
                Late	Fusion	&	96.56	&	90.93	&	97.69	&	95.46	&	0.69	&	4.54	\\	
                Cross	Fusion	&	96.71	&	91.23	&	97.01	&	95.44	&	\textbf{0.59}	&	4.56	\\
                Late+ResNet34	&	93.04		&	87.20		&	93.58		&	92.49		&	1.94	&	7.51	\\	
                PLARD                  & 96.67 & \textbf{93.03} & 97.59          & \textbf{97.75} & 0.74          & 2.25 \\
                \hline
                PSPNet	&	94.30		&	89.04		&	95.60		&	93.84		&	1.31		&	6.96 \\
                DeepLabV3+	&	91.44		&	84.49		&	90.60		&	92.29		&	2.92		&	7.71 \\
                \hline
                SkipcrossNets	&	96.83	$\uparrow$	&	91.03	&	97.79	&	95.89	& 0.66	&	4.11	$\downarrow$	\\
                SkipcrossNets-R	&	\textbf{97.42}	$\uparrow$	&	92.78	$\uparrow$	&	\textbf{97.92}	$\uparrow$	&	96.57		&	0.76	 	&	\textbf{1.97}	$\downarrow$	\\	
                SkipcrossNets-D	&	97.21	$\uparrow$	&	91.47	$\uparrow$	&	97.88	$\uparrow$	&	97.33		&	0.84	 	& 2.19	$\downarrow$	\\	
                \hline \hline
                
                \multirow{2}{*}{Model} & \multicolumn{6}{c}{URBAN UNMARKED ROAD SCENE}   \\
                & MaxF $\uparrow$  & AP  $\uparrow$  & PRE $\uparrow$ & REC $\uparrow$ & FPR $\downarrow$ & FNR $\downarrow$ \\ \hline
                
                Early	Fusion	&	95.60	&	87.72	&	95.13	&	96.08	&	0.78	&	3.92	\\	
                Mid	Fusion	&	96.63	&	89.81	&	97.42	&	95.85	&	0.40	&	4.15	\\	
                Late	Fusion	&	96.85	&	89.64	&	97.24	&	96.46	&	0.43	&	3.54	\\	
                Cross	Fusion	&	96.52	&	89.69	&	97.29	&	95.75	&	0.42	&	4.25	\\	
                Late+ResNet34	&	96.61	&	89.63	&	97.23	&	96.01	&	0.43	&	3.99    \\	
                PLARD                  & 97.34          & \textbf{92.14}   & 97.39          & 97.97   & 0.41          & \textbf{2.03} \\ 
                
                \hline
                PSPNet	&	92.36		&	86.44		&	93.72		&	91.05		&	0.96		&	8.95 \\
                DeepLabV3+	&	83.12		&	69.65		&	75.25		&	92.82		&	4.82		&	7.18 \\
                \hline
                SkipcrossNets	&	96.81	&	89.84	$\uparrow$	&	97.34	&	96.17	&	\textbf{0.40}	$\downarrow$	&	3.83	\\	
                SkipcrossNets-R	&	\textbf{97.76}	$\uparrow$	&	89.78		&	97.44	$\uparrow$	&	97.48	$\uparrow$	&	0.41		&	2.52	$\downarrow$	\\	
                SkipcrossNets-D	&	97.53	$\uparrow$	&	90.78		&	\textbf{97.46}	$\uparrow$	&\textbf{98.81}		$\uparrow$	&	0.48		&	2.19	$\downarrow$	\\	
                \hline
				\end{tabular}
	\end{table}
	
    \subsubsection{Data Preprocessing}
	\label{dataprocessing}
	The inputs to multi-modal network branches on the KITTI dataset include camera images and altitude difference images (ADIs) \cite{PLARD}. The camera data are three-channel RGB images while the ADIs are single-channel images converted from LiDAR point clouds as described in Section \ref{Dataspace}. The ratio of point clouds to images is 0.4\% on the A2D2 dataset, which is highly sparse. Consequently, it is difficult to produce high-quality projection images. As such, we apply K-Nearest Neighbor (KNN) interpolation to complete the point cloud, (i.e., searching the nearest three points for blank pixels and weighting neighboring points based on their distances). This process typically requires large numbers of discrete points to achieve the desired resolution. However, the pixel values of points near the projection image are relatively similar, resulting in poor image quality. Point cloud information typically includes three dimensions: depth, height, and reflection intensity. In this study, a series of experiments are conducted with A2D2 data, in which the height map from point cloud projections is used as input similar to ADIs.

	\subsection{Implementation and Training}
	\label{IAT}
	SkipcrossNets are implemented in PyTorch and trained on an E5-2678v3 CPU running an NVIDIA GTX 1080Ti, with 11 GB of RAM. During training, the batch size is set to 4 and the initial learning rate is 0.001. The learning rate is reduced by a factor of 10 when the validation performance do not improve significantly. The maximum number of epochs is set to 100 and the Adam optimizer is applied in the training process without the use of any pre-trained weights. Road detection tasks are typically considered to be classification problems, so cross-entropy loss is performed in PyTorch. SkipcrossNets are improved by adopting several data augmentation techniques, including multi-scaling, random cropping, distortion of the image brightness, and random removal of road sections.
	
	Increasing the number of stages causes the network layers to become deeper, allowing more abstract features to be extracted at the expense of increased network training costs. Feature fusion also becomes more frequent as the number of blocks increased. However, the number of dense connections also increases exponentially, which make learning more difficult. After several experiments, three stages are selected for feature fusion, with two, three, and three blocks in each successive stage. We conduct an ablation study about first fusion stage in the Section \ref{AS}.
	
	\subsection{Comparisons of Other Fusion Strategies Applied to KITTI}
	\label{KITTI_e}
	Several fusion algorithms are assessed using the KITTI validation set with MaxF as the primary metric. Among these, PLARD \cite{PLARD} uses data spaces and feature adaptation and is currently ranked first for KITTI. SkipscrossNets, the network structure introduced in Section \ref{SK}, is much smaller than PLARD (298 MB) with a network size of only 2.32 MB. Considering the number of layers of the feature extraction network is relatively low (shallow), ResNet34 \cite{resnet} and DenseNet \cite{huang2019convolutional} are selected as backbones for the cross-layer structure in SkipcrossNets-R and the SkipcrossNets-D variant.
	
	Table~\ref{tab:uu} displays test results for various fusion strategies in three independent KITTI scenes. As shown, most multi-modal models produce a MaxF above 95\%, while visual methods perform poorly.
	This is likely because dense KITTI point clouds include additional information unaffected by adverse lighting conditions which provides useful guidance for fusion. The MaxF value is highest for skip-cross fusion in unmarked urban road scenes and marked urban roads. Skip-cross fusion also performs well in marked urban road scenes and multiple marked urban road scenes, achieving better results than other methods. In addition, we observe the FNR of skip-cross fusion is relatively low for multiple fusion strategies. This is primarily due to dense block connection modes adopted in our network structure, which are equivalent to the fusion of information from earlier frames in multiple stages. As a result, SkipcrossNets outperforms other methods in multiple scenes by achieving a lower FNR (i.e. lanes are mistakenly labeled as background), which is critical for autonomous driving.
	
    Table~\ref{tab:uc} displays results for comprehensive urban scenes a combination of the three scenes discussed above. Both PLARD and SkipcrossNets achieve good performance. However, the number of parameters and calculations in SkipcrossNets-D are only 11.4\% and 14.9\% of those in PLARD, respectively. SkipcrossNets are also 33 times faster than PLARD. As hypothesized, the incorporation of the SkipcrossNets structure into ResNet34 and DenseNet leads to additional improvements in the results. Compared with other single-modality methods or multi-modal fusion methods, SkipcrossNets and its variants attain better performance. What's more, SkipcrossNets are also highly adaptable and can be plugged into other backbone structures.

	\begin{table*}[!h]
		\caption{A performance comparison for fusion algorithms applied to urban comprehensive road scenes in the validation set. The ``$-$'' symbol denotes that corresponding information is not reported.}
		\label{tab:uc}
        \def\arraystretch{1.5}
		\begin{center}
			\scalebox{0.78}{
            \rotatebox{90}{
				\begin{tabular}{c|l|llllllccc}
					\hline
					Type & Method & MaxF$\uparrow$ & AP$\uparrow$ & PRE$\uparrow$ & REC$\uparrow$ & FPR$\downarrow$ & FNR$\downarrow$  &PARAMETERS (M)$\downarrow$  & FLOPS (G)$\downarrow$  &FPS$\uparrow$ \\ 
					\hline
					\multirow{7}{*}{single-modal} 
					& PSPNet \cite{PSPNet}  &	92.58		&	86.49		&	93.38		&	91.79		&	1.39		&	8.21  & 2.37 & 5.5  &105.93\\
					&DeeplabV3+ \cite{DeepLabv3+}	&	87.82		&	76.86		&	82.78		&	93.50		&	4.16		&	5.50   &59.34   &166.4  &41.66\\
					&RESA \cite{RESA}	&	94.82		&	88.69		&	94.21		&	94.54		&	1.21		&	2.57   &26.71   &97.27   &29.21\\
					&ENet-SAD \cite{hou2019learning}	&	93.62		&	87.33		&	95.98		&	94.18		&	1.35		&	2.86   &0.81   &2.77   &50.83\\
					&USNet \cite{USNet}	&96.11	&91.71	&95.86	&96.37	&1.36	&3.63  &$-$ & $-$ & $-$\\ 
					&DFM-RTFNet \cite{wang2021dynamic}	&96.58	&92.05	&96.62	&96.93	&1.87	&3.07 & $-$ & $-$ & $-$ \\ 
					&HA-Deeplabv3+ \cite{fan2021learning}	&94.38	&92.12	&94.70	&94.06	&2.40	&5.94  & $-$ & $-$ & $-$ \\ 
					\hline
					\multirow{14}{*}{multi-modal}
					&Early Fusion \cite{wulff2018early} & 95.85 & 88.50 & 95.59 & 96.11 & 0.95 & 3.89 &1.16 &19.23 &114.25\\
					&Mid Fusion \cite{xu2019mid} & 96.55 & 89.95 & 97.19 & 95.93 & 0.59 & 4.07  &2.23 &36.46  &81.30\\
					&Late Fusion \cite{schlosser2016fusing} & 96.63 & 89.89 & 97.12 & 96.15 & 0.61 & 3.85 &2.33 &38.32  &80.64\\
					&Cross Fusion \cite{caltagirone2019lidar} & 96.60 & 90.07 & 97.32 & 95.90 & 0.57 & 4.10 &2.33 &38.32 &54.94\\
					&Late+ResNet34 \cite{resnet}	&94.52	&87.57	&94.57	&94.48	&1.16	&5.52 &45.20  &184.17 &100.14\\
					&CFECA \cite{CFECA}	&85.02	&69.77	&74.98	&96.15	&7.01	&3.85 &26.11  &716.22 &61.1\\
					&PLARD \cite{PLARD}	&97.09	&\textbf{92.50}	&97.63	&97.23	&\textbf{0.51}	&2.65  &76.89 & 588.57 &1.33\\
					&CLCFNet \cite{gu2021cascaded}	&95.65	&89.49	&95.31	&96.00	&2.15	&4.00  & $-$ & $-$ & $-$ \\ 
					&LRDNet+ \cite{LRDNet2022}	&96.10	&92.00	&96.89	&95.32	&1.39	&4.68  & 19.5  & $-$ & $-$ \\ 
					&SNE-RoadSeg \cite{fan2020sne}	&96.42	&91.97	&96.59	&96.26	&1.55	&3.74  & $-$  & $-$ & $-$ \\ 
					&SNE-RoadSeg+ \cite{wang2021sne}	&96.95 &92.08	&96.99	&96.90	&1.37	&3.10  & $-$  & $-$ & $-$ \\ 
					&SkipcrossNets (ours)	&96.85	$\uparrow$ &90.15	$\uparrow$ &97.45	$\uparrow$ &97.14	$\uparrow$ &0.57 $\downarrow$	&2.84$\downarrow$ &2.33 &38.39  &68.24\\
					&SkipcrossNets-R (ours)	&97.38	$\uparrow$ &91.37	$\uparrow$ &97.51	$\uparrow$ &\textbf{97.35}	$\uparrow$ &0.74	&\textbf{2.49}$\downarrow$  &45.67 &335.81 &41.87 \\
					&SkipcrossNets-D (ours)	&\textbf{97.56}	$\uparrow$ &92.38	$\uparrow$ &\textbf{97.76}	$\uparrow$ &97.25	$\uparrow$ &0.68	&3.05$\downarrow$  &8.8 &87.88  &44.05\\
					\hline
			\end{tabular}
			}}
		\end{center}
	\end{table*}
	
	\subsection{Comparisons of Other Fusion Strategies Applied to A2D2}
    \label{A2D2_e}
	Several experiments are conducted using the techniques discussed above, applied to the A2D2 dataset. In the description below, the price of LiDAR equipment corresponds to the density of recorded point clouds. The 128 and 64-line LiDAR exhibit a relatively high price and are not widely used (i.e., most car manufacturers use 16 or 32-line LiDAR as the primary sensor). Thus, The performance of algorithms on sparse point clouds becomes a critical issue.
	
	\begin{table*}[h]
		\caption{A performance comparison for fusion algorithms applied to the A2D2 dataset.}
		\label{tab:A2D2}
		\begin{center}
			\def\arraystretch{1.5}
			\scalebox{0.85}{
				\begin{tabular}{c|l|llllll}
					\hline
					Type & Mehtod & F1$\uparrow$ & PRE$\uparrow$ & REC$\uparrow$ &  MIOU$\uparrow$ & ACC$\uparrow$ & F2$\uparrow$ \\ 
					\hline
					\multirow{5}{*}{single-modal} 
					&PSPNet	\cite{PSPNet} &81.31 &78.49	&81.14		&70.35	&75.94	&80.88\\
					&DeepLabV3+ \cite{DeepLabv3+}	&79.21 &79.41	&83.10		&70.21	&76.47	&72.41\\ 
					&RESA  \cite{RESA}	&82.62 &83.32	&85.17		&74.35	&84.94	&82.88\\
					&ENet-SAD \cite{hou2019learning}	&82.37 &82.41	&86.09		&72.68	&86.47	&80.41\\ 
					&USNet \cite{USNet} &83.26 &83.87	&88.59 &72.23	&88.67	&82.48 \\
					
					\hline
					\multirow{10}{*}{multi-modal} 
					&Early Fusion \cite{wulff2018early} & 83.34 &84.44& 84.24& 73.82 & 92.97& 83.56 \\
					&Mid Fusion \cite{xu2019mid} & 83.14 & 84.31& 83.89 & 73.62 & 92.93& 83.29 \\
					&Late Fusion \cite{schlosser2016fusing} &83.07 & 83.54 &84.28  &73.30 &92.84 &83.49\\
					&Cross Fusion \cite{caltagirone2019lidar} & 82.45 & 83.70& 83.17& 72.41 & 92.57& 82.57\\
					&Late+ResNet34 \cite{resnet} &83.24	&84.49	&83.96		&73.79	&92.96	&83.35\\ 
					&CFECA \cite{CFECA}	&79.52 &75.21	&85.94		&66.77	&89.57	&83.05\\ 
					&PLARD \cite{PLARD}	&83.56 &84.41	&\textbf{88.59}		&67.59	&89.93	&83.86\\
					&LRDNet+ \cite{LRDNet2022} &83.51 &84.26 &87.40		&68.93	&89.48	&83.16 \\
					&SNE-RoadSeg \cite{fan2020sne}  &82.16 &83.34	&87.12		&65.92	&88.13	&82.59 \\
					&SNE-RoadSeg+ \cite{wang2021sne} &82.46 &83.11	&87.53		&66.24	&88.46	&83.05 \\
					&SkipcrossNets (ours) &83.16	&84.84	$\uparrow$ &83.86		&72.88	&92.69	&83.09\\
					&SkipcrossNets-R (ours)	&84.53	$\uparrow$ &\textbf{85.83}	$\uparrow$ &87.23	$\uparrow$  &\textbf{74.21} $\uparrow$	&\textbf{93.13}	$\uparrow$ &83.62 $\uparrow$\\
					&SkipcrossNets-D (ours) &\textbf{84.91}	$\uparrow$	& 84.99	$\uparrow$ &88.23	$\uparrow$  &73.96 $\uparrow$	&92.46 &\textbf{84.24} $\uparrow$\\
					
					\hline
			\end{tabular}}
		\end{center}
	\end{table*}
	
	\begin{figure*}[!htb]
		\centering
		\includegraphics[width=0.9\linewidth]{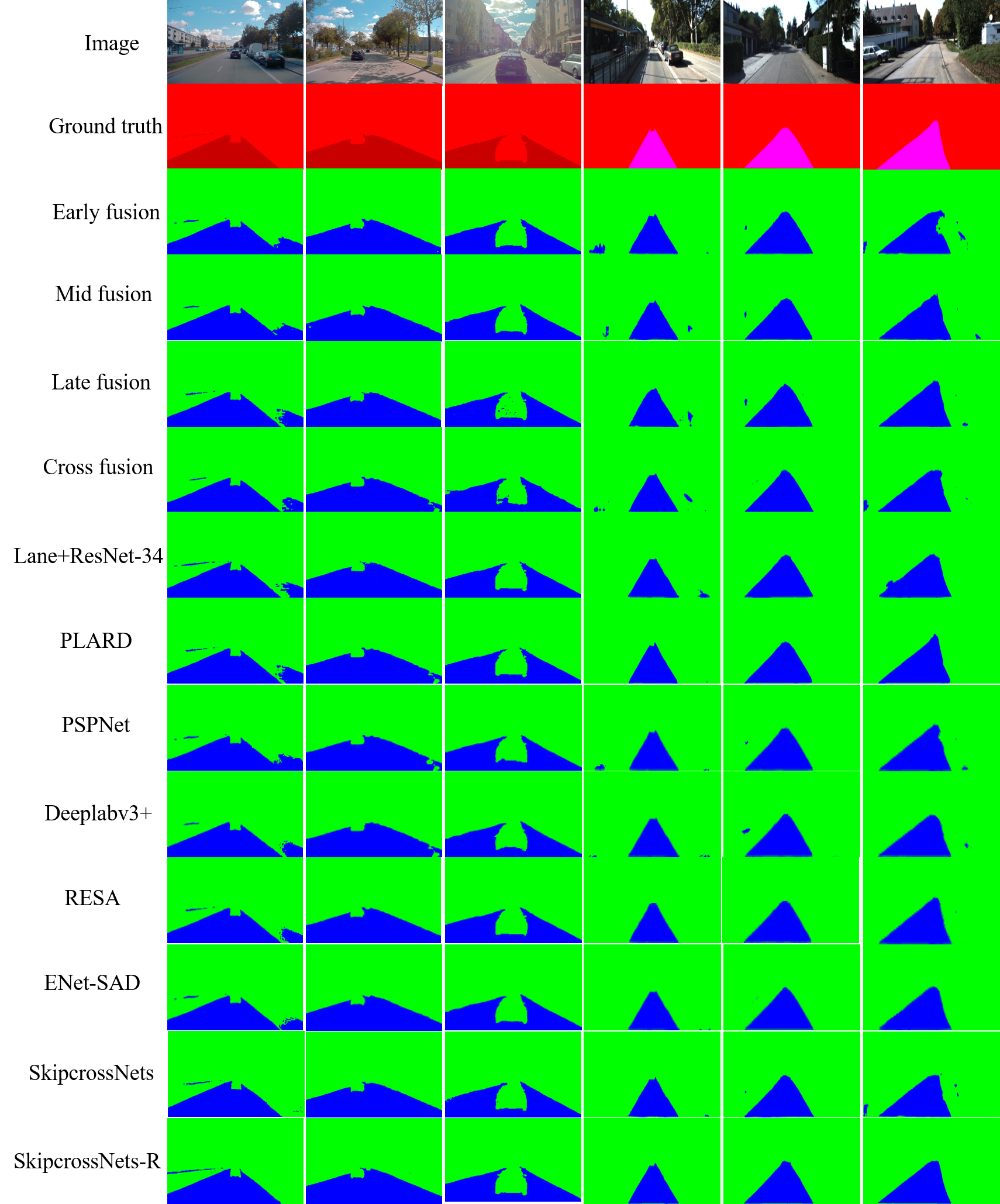}
		\caption{Examples of road segmentation from the KITTI (left three columns) and A2D2 (right three columns) datasets.}
		\label{fig:sumresult}
	\end{figure*}

	Table~\ref{tab:A2D2} compares results for several algorithms with F1 as the primary indicator suggesting that SkipcrossNets achieve the best performance in A2D2 dataset. When A2D2 point cloud data are sparse, the performance of PLARD-based feature extraction is poor. This is because PLARD is modified using a TransNet, and a poor point cloud image does not provide good auxiliary feature effects after adjustment. In contrast, SkipcrossNets enhance sparse point clouds by overlaying adjacent frame features with dense blocks, which is more suitable for these scenes. SkipcrossNets-R also performs well for sparse point cloud data, which is critical for low-cost autonomous driving applications. Compared with other single-modality methods or multi-modal fusion methods, SkipcrossNets and its variants have better performance.
	
	The left three columns and the right three columns in Fig.~\ref{fig:sumresult} show the experimental results for various fusion techniques applied to KITTI and A2D2. The first line shows the original images, the second line shows the ground truths and the other lines show results for various fusion strategies. In the first column, only SkipcrossNets-R is consistent with the ground truth, because it avoids classifying the right lane as a road. In the second column, only cross-fusion and PLARD perform poorly missing a small portion of the road. Most techniques achieve good performance in the third column, but edges in the SkipcrossNets and SkipcrossNets-R results are slightly smoother. In the fourth column, only SkipcrossNet-R results are free from noise. In the fifth column, the bottom five rows exhibit the best performance. In the sixth column, the SkipcrossNets-R results are smoother and free from noise. Overall, SkipcrossNets-R produces lower noise with better performance, making it more suitable for fitting complex models and fusing point clouds with visual images.

	\subsection{Cross-dataset Knowledge}
    \label{CK}
	In this section, we consider improving model detection performance  through cross-dataset transfer learning.
    
	The benefits of knowledge transfer are evaluated experimentally using KITTI and A2D2 data. As shown in Table \ref{tab:cross-dataset}, the model trained on KITTI is directly tested on A2D2, producing an extremely low F1-score of only 1.95\%. In contrast, when the model trained on A2D2 is tested with KITTI, the model learns more difficult scenes (sparse point clouds) and achieves good performance with dense point clouds, producing MAXF and AP values of 78.09\% and 63.3\% respectively. This result is facilitated by varying characteristics of the two data sets, primarily because KITTI uses 64-line LiDAR, while A2D2 uses only one 8-line and two 16-line LiDARs, resulting in significant differences in point cloud density between the two sets. In addition, we fine-tune the model for the corresponding test data and observe the final effects to be slightly less favorable than for a single data set. This confirm that information learned with low point cloud densities exhibited higher robustness for dense point clouds.
    	
    \begin{table}[]
    \caption{The results of cross-dataset knowledge verification experiments with and without fine-tuning.}
    \label{tab:cross-dataset} 
    \centering
    \def\arraystretch{1.6}
    \begin{tabular}{l|c|c|c|llllll}
    \hline
    Method        & Train Dataset & Test Dataset & Fine-tuning & PRE   & REC   & F1    & MIOU  & ACC   & F2    \\ \hline
    SkipcrossNets & KITTI         & A2D2         &             & 54.58 & 1.16  & 1.95  & 1.13  & 76.29 & 1.38  \\
    SkipcrossNets & KITTI         & A2D2         &   \checkmark          & 78.83 & 87.37 & 81.72 & 70.95 & 91.46 & 84.34 \\ 
    SkipcrossNets & A2D2         & A2D2         &   \checkmark          & 84.84 & 83.86 & 83.16 & 72.88 & 92.69 & 83.09 \\
    \hline \hline
    Method        & Train Dataset & Test Dataset & Fine-tuning & MaxF   & AP    & PRE   & REC   & FPR   & FNR   \\ \hline
    SkipcrossNets & A2D2          & KITTI        &             & 78.09 & 63.3  & 67.87 & 91.95 & 9.32  & 8.05  \\
    SkipcrossNets & A2D2          & KITTI        &   \checkmark          & 95.44 & 89.23 & 96.39 & 94.51 & 0.76  & 5.49  \\
    SkipcrossNets & KITTI          & KITTI        &   \checkmark          & 96.85 & 90.15 & 97.45 & 97.14 & 0.57  & 2.84  \\
    \hline
    \end{tabular}
    \end{table}
		\begin{table*}[!htp]
		\centering
		\caption{Ablation study of SkipcrossNets on KITTI dataset. 3 $\sim$ 7th rows show the performance comparison between only using the camera and different factors. ``$+$'' and ``$-$'' respectively represent performance improvement and degradation.}
		\label{tab:ablation}
		\def\arraystretch{1.8}
		\scalebox{0.71}
		{
            \rotatebox{90}{
			\begin{tabular}{ccccccc|cccccc}
				\hline
				\multicolumn{7}{c|}{Method}                                                                    & \multicolumn{6}{c}{Results on KITTI}                         \\ \hline
				Camera & LiDAR & ADIs & Skip-cross & Decoder fusion & ResNet34 & Skip-connection & MaxF/\% $\uparrow$ & AP/\% $\uparrow$ & PRE/\% $\uparrow$ & REC\%  $\uparrow$& FPR/\% $\downarrow$ & FNR/\% $\downarrow$ \\ \hline
				&  \checkmark      &     &                	&                          		&          			&                  	& 94.14   & 87.32 & 94.28  & 93.99 & 1.22   & 6.01  \\
				\checkmark       &        & 		&                    		&                          		&          		&                  	& 95.64   & 88.57 & 95.67  & 95.62 & 0.93   & 4.38  \\ \hline
				\checkmark       &  \checkmark     &  	&			&                          		&          		&          		&-1.24 &-1.50  &-1.66 &-0.84  &+0.36  &+0.84 \\
                \checkmark       &       &  \checkmark	&			&                          		&          		&          		&+0.27 &+0.46  &+0.58 &+0.19  &-0.07  &-0.13 \\
				\checkmark       &       & \checkmark  	&  \checkmark                  &                         		&          		&                  	&+1.11 &+1.29 &+1.42 &+0.80   &-0.31  &-0.80 \\
				\checkmark       &      &   \checkmark	&  \checkmark                  &  \checkmark                        	&          		&                   &+1.21  &+1.58  &+1.78  &+1.52 &-0.36 &-1.54 \\
				\checkmark       &    & \checkmark 	&  \checkmark                  &  \checkmark                        	&  \checkmark        	&           & +1.32 &+1.79  &+1.88 &+1.38  &-0.18  &-1.38 \\
				\checkmark       &      &   \checkmark	&  \checkmark                  &  \checkmark                        	&  \checkmark        	&  \checkmark      &+1.74 & +2.80   &+1.84  &+1.73 &-0.19 &-1.89 \\ \hline
				\multicolumn{7}{c|}{Method}                                                                    & \multicolumn{6}{c}{Results on A2D2}                           \\ \hline
				Camera & LiDAR  & Height map & Skip-cross & Decoder fusion & ResNet34 & Skip-connection & F1/\% $\uparrow$ &  PRE/\% $\uparrow$ & REC/\% $\uparrow$ & MIOU/\% $\uparrow$ & ACC/\% 
 $\uparrow$ & F2/\%  $\uparrow$\\ \hline
				&  \checkmark      &         &           	&                          		&          			&                  	& 77.55  & 81.10   &77.63  & 65.23   & 89.72  & 78.81 \\
				\checkmark       &     &   		&                    		&                          		&          		&                  	& 83.12  & 84.02  &84.33  & 73.48   & 92.86  & 83.12 \\ \hline
				\checkmark       &  \checkmark     & 	&			&                          		&          		&          		 &+0.04 &-0.03  & +0.06 &+0.12  &+0.06  &+0.23 \\
                    \checkmark       &      &  \checkmark 	&			&                          		&          		&          		 &+0.75 &+1.63  &+0.80 &+0.29  &+0.11  &+0.29 \\
				\checkmark       &    &  \checkmark 	&  \checkmark                  &                         		&          		&                  	& +1.18  &+2.70  & +1.28  &+0.46 & +0.23   &+0.43 \\
				\checkmark       &    & \checkmark 	&  \checkmark                  &  \checkmark                        	&          		&             & +1.24 &-0.16 & +0.51 &+0.07  &-0.17 &-0.03 \\
				\checkmark       &    & \checkmark	&  \checkmark                  &  \checkmark                        	&  \checkmark        	&             &+1.27 &-0.02 &+0.04 &+0.03  &+0.01  &+0.18\\
				\checkmark       &       & \checkmark 	&  \checkmark                  &  \checkmark                        	&  \checkmark        	&  \checkmark       & +1.41  &+3.21 &+1.50   &+0.73  &+0.35  &+0.50 \\ \hline
		\end{tabular} 
    }}
	\end{table*}

	\subsection{Ablation Study}
    \label{AS}
	In this section, we explore fusion strategies for multiple scenes. We first consider differences in performance between single-modal and multi-modal strategies, and then discuss the contributions of different factors to the fusion results.
	
	\subsubsection{Comparisons of Different Modalities} 
	The performance of different modalities is verified by training and testing the model with KITTI and A2D2 data. In Table~\ref{tab:ablation}, the first three lines in each sub-table display model performance using LiDAR only, camera only, and two fused information sources, respectively. LiDAR and camera alone produce similar results for KITTI, performing mostly the same function. However, LiDAR data are inferior to camera data for A2D2 because point cloud data in A2D2 are sparser and contain less information. Additionally, the fusion of ADIs and images surpasses the simple fusion strategy of 2D projection of point clouds and images in all metrics on KITTI, and the same holds true for A2D2.
	
	In some cases, the performance of the fusion model is even worse than that of the single-mode model. As seen in the KITTI results of Table~\ref{tab:ablation}, multi-modal fusion is inferior to the use of images only (e.g.,-1.24\%), which is primarily affected by two factors. First, the input suffers from information loss due to the projection of point cloud data from 3D to 2D space. As such, ADIs are used to enhance lane features, as discussed in Section \ref{dataprocessing}. Second, differences between feature spaces make it difficult to ensure that two modal features are at the same level in the fusion process. Thus, we focus on improving the performance of the fusion model and propose a skip-cross fusion strategy. 

\begin{table}[]
\caption{Results of Fusion Stage 1 with the different number of residual blocks.}
\label{tab:fusion_stages}
\def\arraystretch{1.6}
\begin{tabular}{p{5cm} p{0.9cm} p{0.8cm} p{0.8cm} p{0.8cm} p{0.8cm} p{0.9cm}}
\hline
Fusion Stage1                             & MaxF $\uparrow$  & AP $\uparrow$   & PRE $\uparrow$   & REC $\uparrow$  & FPR  $\downarrow$ & FNR $\downarrow$  \\ \hline
2 residual blocks and 1 max pooling layer & \textbf{96.85} & \textbf{90.15} & \textbf{97.43} & \textbf{97.11} & \textbf{0.57} & \textbf{2.84} \\
3 residual blocks and 1 max pooling layer & 95.96 & 88.93 & 96.42 & 96.58 & 0.85 & 3.39 \\ \hline
\end{tabular}
\end{table}

\begin{table}[]
\caption{Results of the skip-cross fusion strategy on different fusion stages and decoders.}
\label{tab:different_fusion_stages}
\def\arraystretch{1.6}
\begin{tabularx}{\textwidth}{XXXXXXXXXX}
\hline
Stage 1 & Stage 2 & Stage 3 & Decoder & MaxF$\uparrow$  & AP$\uparrow$    & PRE$\uparrow$   & REC$\uparrow$   & FPR$\downarrow$  & FNR$\downarrow$  \\ \hline
\centering \checkmark        &         &         &         & 96.03 & 88.97 & 96.74 & 96.81 & 0.71 & 3.29 \\
        & \centering \checkmark       &         &         & 96.35 & 89.64 & 97.21 & 96.96 & 0.62 & 2.97 \\
        &         &\centering \checkmark        &         & 95.76 & 88.52 & 95.84 & 96.31 & 0.83 & 3.44 \\
        &         &         &\centering \checkmark        & 95.73 & 88.60 & 95.79 & 96.25 & 0.88 & 3.36 \\
\centering \checkmark       & \centering \checkmark        & \centering \checkmark        & \centering \checkmark        & \textbf{96.85} & \textbf{90.15} & \textbf{97.45} & \textbf{97.14} & \textbf{0.57} & \textbf{2.84} \\ \hline
\end{tabularx}
\end{table}
	
	\subsubsection{Comparisons of Different Factors}
	To further explore optimal fusion model performance, we conduct experiments using four different strategies, skip-cross fusion, decoder fusion (with skip-cross fusion), ResNet34, and skip connections.
	
	As shown in Table~\ref{tab:ablation}, skip-cross fusion increases MaxF by 1.11\% and F1 by 1.18\% in the case of KITTI and A2D2, respectively. Skip-cross fusion can adaptively adjust the weights of two data sources producing better fusion results than other strategies. 
    The Skip-cross fusion also results in improvements for KITTI and A2D2 when the decoder is frequently used to adjust the data source, restoring image details and resolution. Since a deeper model implies better nonlinear expression capabilities, more complex transformations can be learned, and more complex feature inputs can be fitted as a result. Thus, ResNet34 is employed as the backbone to deepen the network and implement skip-cross fusion thereby increasing MaxF by 1.32\% and F1 by 1.27\%. Furthermore, skip connections are added to directly transfer feature maps from the encoder to the decoder, achieving better performance. 
    This skip-cross fusion strategy is highly generalizable, as confirmed by its integration into a ResNet34, and can be similarly extended to other network structures, such as DenseNet.

    Table \ref{tab:fusion_stages} shows the performance comparison of Fusion Stage 1 with the different number of residual blocks. It is clear that the model with 2 residual blocks surpasses the one with 3 residual blocks by 0.89\% in terms of MaxF. This is attributed to the fact that the first residual block, situated in the early stages of the convolutional neural network, fails to capture sufficiently rich high-level semantic information, resulting in a lack of comprehensive understanding of complex features. In Table \ref{tab:different_fusion_stages}, we show the ablation experiments of the skip-cross fusion strategy at different fusion stages and decoders. Due to the fact that deep convolutional layers extract more advanced semantic features, while shallow convolutional layers capture rich detailed information, solely incorporating the skip-cross fusion strategy in each Fusion Stage and Decoder does not fully leverage both semantic features and abundant detailed information, leading to a certain performance decline.

	\section{Conclusion}
	A new skip-cross fusion network structure is proposed, which combines Altitude Difference Images from LiDAR point clouds with camera images for improving road detection. 
    The proposed skip cross-fusion adaptively adjusts fusion parameters during the training process to determine the optimal fusion timing and level, thereby fully promoting the fusion and reuse of features at different levels.
	In the presented experiments, SkipcrossNets provide several performance advantages compared with existing fusion strategies, achieving promising performance for the KITTI road set and A2D2 datasets. In addition, SkipcrossNets are small (only 2.32 MB) and capable of meeting real-time inference speed, reaching 68.24 FPS. The SkipcrossNets-R and SkipcrossNets-D variants are also proposed to improve detection accuracy, exhibiting deeper model layers and offering better performance.
	In the future, we will further study the role of different skip-cross connections in a network, removing redundant connections to reduce the number of parameters and computations.

\section{Declarations}
\textbf{Funding}
This work was supported by the National High Technology Research and Development Program of China under Grant No. 2022YFD2002305, and the National Natural Science Foundation of China under Grant No. 62273198.

\noindent \textbf{Conflict of interest}
The authors have no competing interests to declare
that are relevant to the content of this article.


\bibliography{myref}


\begin{thebibliography}{67}
\ifx \bisbn   \undefined \def \bisbn  #1{ISBN #1}\fi
\ifx \binits  \undefined \def \binits#1{#1}\fi
\ifx \bauthor  \undefined \def \bauthor#1{#1}\fi
\ifx \batitle  \undefined \def \batitle#1{#1}\fi
\ifx \bjtitle  \undefined \def \bjtitle#1{#1}\fi
\ifx \bvolume  \undefined \def \bvolume#1{\textbf{#1}}\fi
\ifx \byear  \undefined \def \byear#1{#1}\fi
\ifx \bissue  \undefined \def \bissue#1{#1}\fi
\ifx \bfpage  \undefined \def \bfpage#1{#1}\fi
\ifx \blpage  \undefined \def \blpage #1{#1}\fi
\ifx \burl  \undefined \def \burl#1{\textsf{#1}}\fi
\ifx \doiurl  \undefined \def \doiurl#1{\url{https://doi.org/#1}}\fi
\ifx \betal  \undefined \def \betal{\textit{et al.}}\fi
\ifx \binstitute  \undefined \def \binstitute#1{#1}\fi
\ifx \binstitutionaled  \undefined \def \binstitutionaled#1{#1}\fi
\ifx \bctitle  \undefined \def \bctitle#1{#1}\fi
\ifx \beditor  \undefined \def \beditor#1{#1}\fi
\ifx \bpublisher  \undefined \def \bpublisher#1{#1}\fi
\ifx \bbtitle  \undefined \def \bbtitle#1{#1}\fi
\ifx \bedition  \undefined \def \bedition#1{#1}\fi
\ifx \bseriesno  \undefined \def \bseriesno#1{#1}\fi
\ifx \blocation  \undefined \def \blocation#1{#1}\fi
\ifx \bsertitle  \undefined \def \bsertitle#1{#1}\fi
\ifx \bsnm \undefined \def \bsnm#1{#1}\fi
\ifx \bsuffix \undefined \def \bsuffix#1{#1}\fi
\ifx \bparticle \undefined \def \bparticle#1{#1}\fi
\ifx \barticle \undefined \def \barticle#1{#1}\fi
\bibcommenthead
\ifx \bconfdate \undefined \def \bconfdate #1{#1}\fi
\ifx \botherref \undefined \def \botherref #1{#1}\fi
\ifx \url \undefined \def \url#1{\textsf{#1}}\fi
\ifx \bchapter \undefined \def \bchapter#1{#1}\fi
\ifx \bbook \undefined \def \bbook#1{#1}\fi
\ifx \bcomment \undefined \def \bcomment#1{#1}\fi
\ifx \oauthor \undefined \def \oauthor#1{#1}\fi
\ifx \citeauthoryear \undefined \def \citeauthoryear#1{#1}\fi
\ifx \endbibitem  \undefined \def \endbibitem {}\fi
\ifx \bconflocation  \undefined \def \bconflocation#1{#1}\fi
\ifx \arxivurl  \undefined \def \arxivurl#1{\textsf{#1}}\fi
\csname PreBibitemsHook\endcsname

\bibitem{kang2011multiband}
\begin{barticle}
\bauthor{\bsnm{Kang}, \binits{Y.}},
\bauthor{\bsnm{Yamaguchi}, \binits{K.}},
\bauthor{\bsnm{Naito}, \binits{T.}},
\bauthor{\bsnm{Ninomiya}, \binits{Y.}}:
\batitle{Multiband image segmentation and object recognition for understanding road scenes}.
\bjtitle{IEEE Transactions on Intelligent Transportation Systems}
\bvolume{12}(\bissue{4}),
\bfpage{1423}--\blpage{1433}
(\byear{2011})
\end{barticle}
\endbibitem

\bibitem{zhang2021multi}
\begin{bchapter}
\bauthor{\bsnm{Zhang}, \binits{X.}},
\bauthor{\bsnm{Gong}, \binits{Y.}},
\bauthor{\bsnm{Li}, \binits{Z.}},
\bauthor{\bsnm{Liu}, \binits{X.}},
\bauthor{\bsnm{Pan}, \binits{S.}},
\bauthor{\bsnm{Li}, \binits{J.}}:
\bctitle{Multi-modal attention guided real-time lane detection}.
In: \bbtitle{2021 6th IEEE International Conference on Advanced Robotics and Mechatronics (ICARM)},
pp. \bfpage{146}--\blpage{153}
(\byear{2021}).
\bcomment{IEEE}
\end{bchapter}
\endbibitem

\bibitem{gong2024tclanenet}
\begin{botherref}
\oauthor{\bsnm{Gong}, \binits{Y.}},
\oauthor{\bsnm{Jiang}, \binits{X.}},
\oauthor{\bsnm{Wang}, \binits{L.}},
\oauthor{\bsnm{Xu}, \binits{L.}},
\oauthor{\bsnm{Lu}, \binits{J.}},
\oauthor{\bsnm{Liu}, \binits{H.}},
\oauthor{\bsnm{Lin}, \binits{L.}},
\oauthor{\bsnm{Zhang}, \binits{X.}}:
Tclanenet: Task-conditioned lane detection network driven by vibration information.
IEEE Transactions on Intelligent Vehicles
(2024)
\end{botherref}
\endbibitem

\bibitem{wan2024adnet}
\begin{barticle}
\bauthor{\bsnm{Wan}, \binits{B.}},
\bauthor{\bsnm{Zhou}, \binits{X.}},
\bauthor{\bsnm{Sun}, \binits{Y.}},
\bauthor{\bsnm{Wang}, \binits{T.}},
\bauthor{\bsnm{Wang}, \binits{S.}},
\bauthor{\bsnm{Yin}, \binits{H.}},
\bauthor{\bsnm{Yan}, \binits{C.}}, \betal:
\batitle{Adnet: Anti-noise dual-branch network for road defect detection}.
\bjtitle{Engineering Applications of Artificial Intelligence}
\bvolume{132},
\bfpage{107963}
(\byear{2024})
\end{barticle}
\endbibitem

\bibitem{bandara2022spin}
\begin{bchapter}
\bauthor{\bsnm{Bandara}, \binits{W.G.C.}},
\bauthor{\bsnm{Valanarasu}, \binits{J.M.J.}},
\bauthor{\bsnm{Patel}, \binits{V.M.}}:
\bctitle{Spin road mapper: Extracting roads from aerial images via spatial and interaction space graph reasoning for autonomous driving}.
In: \bbtitle{2022 International Conference on Robotics and Automation (ICRA)},
pp. \bfpage{343}--\blpage{350}
(\byear{2022}).
\bcomment{IEEE}
\end{bchapter}
\endbibitem

\bibitem{chang2022fast}
\begin{bchapter}
\bauthor{\bsnm{Chang}, \binits{Y.}},
\bauthor{\bsnm{Xue}, \binits{F.}},
\bauthor{\bsnm{Sheng}, \binits{F.}},
\bauthor{\bsnm{Liang}, \binits{W.}},
\bauthor{\bsnm{Ming}, \binits{A.}}:
\bctitle{Fast road segmentation via uncertainty-aware symmetric network}.
In: \bbtitle{2022 International Conference on Robotics and Automation (ICRA)},
pp. \bfpage{11124}--\blpage{11130}
(\byear{2022}).
\bcomment{IEEE}
\end{bchapter}
\endbibitem

\bibitem{alvarez2012road}
\begin{bchapter}
\bauthor{\bsnm{Alvarez}, \binits{J.M.}},
\bauthor{\bsnm{Gevers}, \binits{T.}},
\bauthor{\bsnm{LeCun}, \binits{Y.}},
\bauthor{\bsnm{Lopez}, \binits{A.M.}}:
\bctitle{Road scene segmentation from a single image}.
In: \bbtitle{European Conference on Computer Vision},
pp. \bfpage{376}--\blpage{389}
(\byear{2012}).
\bcomment{Springer}
\end{bchapter}
\endbibitem

\bibitem{zhou2024decoupling}
\begin{botherref}
\oauthor{\bsnm{Zhou}, \binits{X.}},
\oauthor{\bsnm{Wu}, \binits{Z.}},
\oauthor{\bsnm{Cong}, \binits{R.}}:
Decoupling and integration network for camouflaged object detection.
IEEE Transactions on Multimedia
(2024)
\end{botherref}
\endbibitem

\bibitem{wan2023lfrnet}
\begin{barticle}
\bauthor{\bsnm{Wan}, \binits{B.}},
\bauthor{\bsnm{Zhou}, \binits{X.}},
\bauthor{\bsnm{Zheng}, \binits{B.}},
\bauthor{\bsnm{Yin}, \binits{H.}},
\bauthor{\bsnm{Zhu}, \binits{Z.}},
\bauthor{\bsnm{Wang}, \binits{H.}},
\bauthor{\bsnm{Sun}, \binits{Y.}},
\bauthor{\bsnm{Zhang}, \binits{J.}},
\bauthor{\bsnm{Yan}, \binits{C.}}:
\batitle{Lfrnet: Localizing, focus, and refinement network for salient object detection of surface defects}.
\bjtitle{IEEE Transactions on Instrumentation and Measurement}
\bvolume{72},
\bfpage{1}--\blpage{12}
(\byear{2023})
\end{barticle}
\endbibitem

\bibitem{PLARD}
\begin{barticle}
\bauthor{\bsnm{Chen}, \binits{Z.}},
\bauthor{\bsnm{Zhang}, \binits{J.}},
\bauthor{\bsnm{Tao}, \binits{D.}}:
\batitle{Progressive lidar adaptation for road detection}.
\bjtitle{IEEE/CAA Journal of Automatica Sinica}
\bvolume{6}(\bissue{3}),
\bfpage{693}--\blpage{702}
(\byear{2019})
\end{barticle}
\endbibitem

\bibitem{tomas2016forecasting}
\begin{barticle}
\bauthor{\bsnm{Tom{\'a}s}, \binits{V.R.}},
\bauthor{\bsnm{Pla-Castells}, \binits{M.}},
\bauthor{\bsnm{Mart{\'\i}nez}, \binits{J.J.}},
\bauthor{\bsnm{Mart{\'\i}nez}, \binits{J.}}:
\batitle{Forecasting adverse weather situations in the road network}.
\bjtitle{IEEE Transactions on Intelligent Transportation Systems}
\bvolume{17}(\bissue{8}),
\bfpage{2334}--\blpage{2343}
(\byear{2016})
\end{barticle}
\endbibitem

\bibitem{zhang2023multi}
\begin{botherref}
\oauthor{\bsnm{Zhang}, \binits{X.}},
\oauthor{\bsnm{Gong}, \binits{Y.}},
\oauthor{\bsnm{Lu}, \binits{J.}},
\oauthor{\bsnm{Wu}, \binits{J.}},
\oauthor{\bsnm{Li}, \binits{Z.}},
\oauthor{\bsnm{Jin}, \binits{D.}},
\oauthor{\bsnm{Li}, \binits{J.}}:
Multi-modal fusion technology based on vehicle information: A survey.
IEEE Transactions on Intelligent Vehicles
(2023)
\end{botherref}
\endbibitem

\bibitem{gong2023sifdrivenet}
\begin{botherref}
\oauthor{\bsnm{Gong}, \binits{Y.}},
\oauthor{\bsnm{Lu}, \binits{J.}},
\oauthor{\bsnm{Liu}, \binits{W.}},
\oauthor{\bsnm{Li}, \binits{Z.}},
\oauthor{\bsnm{Jiang}, \binits{X.}},
\oauthor{\bsnm{Gao}, \binits{X.}},
\oauthor{\bsnm{Wu}, \binits{X.}}:
Sifdrivenet: Speed and image fusion for driving behavior classification network.
IEEE Transactions on Computational Social Systems
(2023)
\end{botherref}
\endbibitem

\bibitem{feng2020deep}
\begin{barticle}
\bauthor{\bsnm{Feng}, \binits{D.}},
\bauthor{\bsnm{Haase-Sch{\"u}tz}, \binits{C.}},
\bauthor{\bsnm{Rosenbaum}, \binits{L.}},
\bauthor{\bsnm{Hertlein}, \binits{H.}},
\bauthor{\bsnm{Glaeser}, \binits{C.}},
\bauthor{\bsnm{Timm}, \binits{F.}},
\bauthor{\bsnm{Wiesbeck}, \binits{W.}},
\bauthor{\bsnm{Dietmayer}, \binits{K.}}:
\batitle{Deep multi-modal object detection and semantic segmentation for autonomous driving: Datasets, methods, and challenges}.
\bjtitle{IEEE Transactions on Intelligent Transportation Systems}
\bvolume{22}(\bissue{3}),
\bfpage{1341}--\blpage{1360}
(\byear{2020})
\end{barticle}
\endbibitem

\bibitem{gong2023feature}
\begin{botherref}
\oauthor{\bsnm{Gong}, \binits{Y.}},
\oauthor{\bsnm{Wang}, \binits{L.}},
\oauthor{\bsnm{Xu}, \binits{L.}}:
A feature aggregation network for multispectral pedestrian detection.
Applied Intelligence,
1--15
(2023)
\end{botherref}
\endbibitem

\bibitem{wulff2018early}
\begin{bchapter}
\bauthor{\bsnm{Wulff}, \binits{F.}},
\bauthor{\bsnm{Sch{\"a}ufele}, \binits{B.}},
\bauthor{\bsnm{Sawade}, \binits{O.}},
\bauthor{\bsnm{Becker}, \binits{D.}},
\bauthor{\bsnm{Henke}, \binits{B.}},
\bauthor{\bsnm{Radusch}, \binits{I.}}:
\bctitle{Early fusion of camera and lidar for robust road detection based on u-net fcn}.
In: \bbtitle{2018 IEEE Intelligent Vehicles Symposium (IV)},
pp. \bfpage{1426}--\blpage{1431}
(\byear{2018}).
\bcomment{IEEE}
\end{bchapter}
\endbibitem

\bibitem{xu2019mid}
\begin{bchapter}
\bauthor{\bsnm{Xu}, \binits{B.}},
\bauthor{\bsnm{Li}, \binits{W.}},
\bauthor{\bsnm{Tzoumanikas}, \binits{D.}},
\bauthor{\bsnm{Bloesch}, \binits{M.}},
\bauthor{\bsnm{Davison}, \binits{A.}},
\bauthor{\bsnm{Leutenegger}, \binits{S.}}:
\bctitle{Mid-fusion: Octree-based object-level multi-instance dynamic slam}.
In: \bbtitle{2019 International Conference on Robotics and Automation (ICRA)},
pp. \bfpage{5231}--\blpage{5237}
(\byear{2019}).
\bcomment{IEEE}
\end{bchapter}
\endbibitem

\bibitem{schlosser2016fusing}
\begin{bchapter}
\bauthor{\bsnm{Schlosser}, \binits{J.}},
\bauthor{\bsnm{Chow}, \binits{C.K.}},
\bauthor{\bsnm{Kira}, \binits{Z.}}:
\bctitle{Fusing lidar and images for pedestrian detection using convolutional neural networks}.
In: \bbtitle{2016 IEEE International Conference on Robotics and Automation (ICRA)},
pp. \bfpage{2198}--\blpage{2205}
(\byear{2016}).
\bcomment{IEEE}
\end{bchapter}
\endbibitem

\bibitem{liu2024glmdrivenet}
\begin{barticle}
\bauthor{\bsnm{Liu}, \binits{W.}},
\bauthor{\bsnm{Gong}, \binits{Y.}},
\bauthor{\bsnm{Zhang}, \binits{G.}},
\bauthor{\bsnm{Lu}, \binits{J.}},
\bauthor{\bsnm{Zhou}, \binits{Y.}},
\bauthor{\bsnm{Liao}, \binits{J.}}:
\batitle{Glmdrivenet: Global--local multimodal fusion driving behavior classification network}.
\bjtitle{Engineering Applications of Artificial Intelligence}
\bvolume{129},
\bfpage{107575}
(\byear{2024})
\end{barticle}
\endbibitem

\bibitem{resnet}
\begin{bchapter}
\bauthor{\bsnm{He}, \binits{K.}},
\bauthor{\bsnm{Zhang}, \binits{X.}},
\bauthor{\bsnm{Ren}, \binits{S.}},
\bauthor{\bsnm{Sun}, \binits{J.}}:
\bctitle{Deep residual learning for image recognition}.
In: \bbtitle{Proceedings of the IEEE Conference on Computer Vision and Pattern Recognition},
pp. \bfpage{770}--\blpage{778}
(\byear{2016})
\end{bchapter}
\endbibitem

\bibitem{huang2019convolutional}
\begin{botherref}
\oauthor{\bsnm{Huang}, \binits{G.}},
\oauthor{\bsnm{Liu}, \binits{Z.}},
\oauthor{\bsnm{Pleiss}, \binits{G.}},
\oauthor{\bsnm{Van Der~Maaten}, \binits{L.}},
\oauthor{\bsnm{Weinberger}, \binits{K.}}:
Convolutional networks with dense connectivity.
IEEE transactions on pattern analysis and machine intelligence
(2019)
\end{botherref}
\endbibitem

\bibitem{caltagirone2019lidar}
\begin{barticle}
\bauthor{\bsnm{Caltagirone}, \binits{L.}},
\bauthor{\bsnm{Bellone}, \binits{M.}},
\bauthor{\bsnm{Svensson}, \binits{L.}},
\bauthor{\bsnm{Wahde}, \binits{M.}}:
\batitle{Lidar-camera fusion for road detection using fully convolutional neural networks}.
\bjtitle{Robotics and Autonomous Systems}
\bvolume{111},
\bfpage{125}--\blpage{131}
(\byear{2019})
\end{barticle}
\endbibitem

\bibitem{PSPNet}
\begin{bchapter}
\bauthor{\bsnm{Zhao}, \binits{H.}},
\bauthor{\bsnm{Shi}, \binits{J.}},
\bauthor{\bsnm{Qi}, \binits{X.}},
\bauthor{\bsnm{Wang}, \binits{X.}},
\bauthor{\bsnm{Jia}, \binits{J.}}:
\bctitle{Pyramid scene parsing network}.
In: \bbtitle{Proceedings of the IEEE Conference on Computer Vision and Pattern Recognition},
pp. \bfpage{2881}--\blpage{2890}
(\byear{2017})
\end{bchapter}
\endbibitem

\bibitem{DeepLabv3+}
\begin{bchapter}
\bauthor{\bsnm{Chen}, \binits{L.-C.}},
\bauthor{\bsnm{Zhu}, \binits{Y.}},
\bauthor{\bsnm{Papandreou}, \binits{G.}},
\bauthor{\bsnm{Schroff}, \binits{F.}},
\bauthor{\bsnm{Adam}, \binits{H.}}:
\bctitle{Encoder-decoder with atrous separable convolution for semantic image segmentation}.
In: \bbtitle{Proceedings of the European Conference on Computer Vision (ECCV)},
pp. \bfpage{801}--\blpage{818}
(\byear{2018})
\end{bchapter}
\endbibitem

\bibitem{geiger2012we}
\begin{bchapter}
\bauthor{\bsnm{Geiger}, \binits{A.}},
\bauthor{\bsnm{Lenz}, \binits{P.}},
\bauthor{\bsnm{Urtasun}, \binits{R.}}:
\bctitle{Are we ready for autonomous driving? the kitti vision benchmark suite}.
In: \bbtitle{2012 IEEE Conference on Computer Vision and Pattern Recognition},
pp. \bfpage{3354}--\blpage{3361}
(\byear{2012}).
\bcomment{IEEE}
\end{bchapter}
\endbibitem

\bibitem{baltruvsaitis2018multimodal}
\begin{barticle}
\bauthor{\bsnm{Baltru{\v{s}}aitis}, \binits{T.}},
\bauthor{\bsnm{Ahuja}, \binits{C.}},
\bauthor{\bsnm{Morency}, \binits{L.-P.}}:
\batitle{Multimodal machine learning: A survey and taxonomy}.
\bjtitle{IEEE transactions on pattern analysis and machine intelligence}
\bvolume{41}(\bissue{2}),
\bfpage{423}--\blpage{443}
(\byear{2018})
\end{barticle}
\endbibitem

\bibitem{sun2006hsi}
\begin{bchapter}
\bauthor{\bsnm{Sun}, \binits{T.-Y.}},
\bauthor{\bsnm{Tsai}, \binits{S.-J.}},
\bauthor{\bsnm{Chan}, \binits{V.}}:
\bctitle{Hsi color model based lane-marking detection}.
In: \bbtitle{2006 IEEE Intelligent Transportation Systems Conference},
pp. \bfpage{1168}--\blpage{1172}
(\byear{2006}).
\bcomment{IEEE}
\end{bchapter}
\endbibitem

\bibitem{yu1997lane}
\begin{bchapter}
\bauthor{\bsnm{Yu}, \binits{B.}},
\bauthor{\bsnm{Jain}, \binits{A.K.}}:
\bctitle{Lane boundary detection using a multiresolution hough transform}.
In: \bbtitle{Proceedings of International Conference on Image Processing},
vol. \bseriesno{2},
pp. \bfpage{748}--\blpage{751}
(\byear{1997}).
\bcomment{IEEE}
\end{bchapter}
\endbibitem

\bibitem{wang2000lane}
\begin{barticle}
\bauthor{\bsnm{Wang}, \binits{Y.}},
\bauthor{\bsnm{Shen}, \binits{D.}},
\bauthor{\bsnm{Teoh}, \binits{E.K.}}:
\batitle{Lane detection using spline model}.
\bjtitle{Pattern Recognition Letters}
\bvolume{21}(\bissue{8}),
\bfpage{677}--\blpage{689}
(\byear{2000})
\end{barticle}
\endbibitem

\bibitem{laptev2000automatic}
\begin{barticle}
\bauthor{\bsnm{Laptev}, \binits{I.}},
\bauthor{\bsnm{Mayer}, \binits{H.}},
\bauthor{\bsnm{Lindeberg}, \binits{T.}},
\bauthor{\bsnm{Eckstein}, \binits{W.}},
\bauthor{\bsnm{Steger}, \binits{C.}},
\bauthor{\bsnm{Baumgartner}, \binits{A.}}:
\batitle{Automatic extraction of roads from aerial images based on scale space and snakes}.
\bjtitle{Machine Vision and Applications}
\bvolume{12}(\bissue{1}),
\bfpage{23}--\blpage{31}
(\byear{2000})
\end{barticle}
\endbibitem

\bibitem{wegner2013higher}
\begin{bchapter}
\bauthor{\bsnm{Wegner}, \binits{J.D.}},
\bauthor{\bsnm{Montoya-Zegarra}, \binits{J.A.}},
\bauthor{\bsnm{Schindler}, \binits{K.}}:
\bctitle{A higher-order crf model for road network extraction}.
In: \bbtitle{Proceedings of the IEEE Conference on Computer Vision and Pattern Recognition},
pp. \bfpage{1698}--\blpage{1705}
(\byear{2013})
\end{bchapter}
\endbibitem

\bibitem{chai2013recovering}
\begin{bchapter}
\bauthor{\bsnm{Chai}, \binits{D.}},
\bauthor{\bsnm{Forstner}, \binits{W.}},
\bauthor{\bsnm{Lafarge}, \binits{F.}}:
\bctitle{Recovering line-networks in images by junction-point processes}.
In: \bbtitle{Proceedings of the IEEE Conference on Computer Vision and Pattern Recognition},
pp. \bfpage{1894}--\blpage{1901}
(\byear{2013})
\end{bchapter}
\endbibitem

\bibitem{stoica2004gibbs}
\begin{barticle}
\bauthor{\bsnm{Stoica}, \binits{R.}},
\bauthor{\bsnm{Descombes}, \binits{X.}},
\bauthor{\bsnm{Zerubia}, \binits{J.}}:
\batitle{A gibbs point process for road extraction from remotely sensed images}.
\bjtitle{International Journal of Computer Vision}
\bvolume{57}(\bissue{2}),
\bfpage{121}--\blpage{136}
(\byear{2004})
\end{barticle}
\endbibitem

\bibitem{wang2015adaptive}
\begin{barticle}
\bauthor{\bsnm{Wang}, \binits{Q.}},
\bauthor{\bsnm{Fang}, \binits{J.}},
\bauthor{\bsnm{Yuan}, \binits{Y.}}:
\batitle{Adaptive road detection via context-aware label transfer}.
\bjtitle{Neurocomputing}
\bvolume{158},
\bfpage{174}--\blpage{183}
(\byear{2015})
\end{barticle}
\endbibitem

\bibitem{qi2018dynamic}
\begin{barticle}
\bauthor{\bsnm{Qi}, \binits{L.}},
\bauthor{\bsnm{Zhou}, \binits{M.}},
\bauthor{\bsnm{Luan}, \binits{W.}}:
\batitle{A dynamic road incident information delivery strategy to reduce urban traffic congestion}.
\bjtitle{IEEE/CAA Journal of Automatica Sinica}
\bvolume{5}(\bissue{5}),
\bfpage{934}--\blpage{945}
(\byear{2018})
\end{barticle}
\endbibitem

\bibitem{chen2018parallel}
\begin{barticle}
\bauthor{\bsnm{Chen}, \binits{L.}},
\bauthor{\bsnm{Hu}, \binits{X.}},
\bauthor{\bsnm{Tian}, \binits{W.}},
\bauthor{\bsnm{Wang}, \binits{H.}},
\bauthor{\bsnm{Cao}, \binits{D.}},
\bauthor{\bsnm{Wang}, \binits{F.-Y.}}:
\batitle{Parallel planning: A new motion planning framework for autonomous driving}.
\bjtitle{IEEE/CAA Journal of Automatica Sinica}
\bvolume{6}(\bissue{1}),
\bfpage{236}--\blpage{246}
(\byear{2018})
\end{barticle}
\endbibitem

\bibitem{kong2009vanishing}
\begin{bchapter}
\bauthor{\bsnm{Kong}, \binits{H.}},
\bauthor{\bsnm{Audibert}, \binits{J.-Y.}},
\bauthor{\bsnm{Ponce}, \binits{J.}}:
\bctitle{Vanishing point detection for road detection}.
In: \bbtitle{2009 Ieee Conference on Computer Vision and Pattern Recognition},
pp. \bfpage{96}--\blpage{103}
(\byear{2009}).
\bcomment{IEEE}
\end{bchapter}
\endbibitem

\bibitem{zhang2022openmpd}
\begin{barticle}
\bauthor{\bsnm{Zhang}, \binits{X.}},
\bauthor{\bsnm{Li}, \binits{Z.}},
\bauthor{\bsnm{Gong}, \binits{Y.}},
\bauthor{\bsnm{Jin}, \binits{D.}},
\bauthor{\bsnm{Li}, \binits{J.}},
\bauthor{\bsnm{Wang}, \binits{L.}},
\bauthor{\bsnm{Zhu}, \binits{Y.}},
\bauthor{\bsnm{Liu}, \binits{H.}}:
\batitle{Openmpd: An open multimodal perception dataset for autonomous driving}.
\bjtitle{IEEE Transactions on Vehicular Technology}
\bvolume{71}(\bissue{3}),
\bfpage{2437}--\blpage{2447}
(\byear{2022})
\end{barticle}
\endbibitem

\bibitem{chen2017generic}
\begin{bchapter}
\bauthor{\bsnm{Chen}, \binits{Z.}},
\bauthor{\bsnm{Li}, \binits{J.}},
\bauthor{\bsnm{Chen}, \binits{Z.}},
\bauthor{\bsnm{You}, \binits{X.}}:
\bctitle{Generic pixel level object tracker using bi-channel fully convolutional network}.
In: \bbtitle{International Conference on Neural Information Processing},
pp. \bfpage{666}--\blpage{676}
(\byear{2017}).
\bcomment{Springer}
\end{bchapter}
\endbibitem

\bibitem{zhang2023oblique}
\begin{botherref}
\oauthor{\bsnm{Zhang}, \binits{X.}},
\oauthor{\bsnm{Gong}, \binits{Y.}},
\oauthor{\bsnm{Lu}, \binits{J.}},
\oauthor{\bsnm{Li}, \binits{Z.}},
\oauthor{\bsnm{Li}, \binits{S.}},
\oauthor{\bsnm{Wang}, \binits{S.}},
\oauthor{\bsnm{Liu}, \binits{W.}},
\oauthor{\bsnm{Wang}, \binits{L.}},
\oauthor{\bsnm{Li}, \binits{J.}}:
Oblique convolution: A novel convolution idea for redefining lane detection.
IEEE Transactions on Intelligent Vehicles
(2023)
\end{botherref}
\endbibitem

\bibitem{teichmann2018multinet}
\begin{bchapter}
\bauthor{\bsnm{Teichmann}, \binits{M.}},
\bauthor{\bsnm{Weber}, \binits{M.}},
\bauthor{\bsnm{Zoellner}, \binits{M.}},
\bauthor{\bsnm{Cipolla}, \binits{R.}},
\bauthor{\bsnm{Urtasun}, \binits{R.}}:
\bctitle{Multinet: Real-time joint semantic reasoning for autonomous driving}.
In: \bbtitle{2018 IEEE Intelligent Vehicles Symposium (IV)},
pp. \bfpage{1013}--\blpage{1020}
(\byear{2018}).
\bcomment{IEEE}
\end{bchapter}
\endbibitem

\bibitem{he2016accurate}
\begin{bchapter}
\bauthor{\bsnm{He}, \binits{B.}},
\bauthor{\bsnm{Ai}, \binits{R.}},
\bauthor{\bsnm{Yan}, \binits{Y.}},
\bauthor{\bsnm{Lang}, \binits{X.}}:
\bctitle{Accurate and robust lane detection based on dual-view convolutional neutral network}.
In: \bbtitle{2016 IEEE Intelligent Vehicles Symposium (IV)},
pp. \bfpage{1041}--\blpage{1046}
(\byear{2016}).
\bcomment{IEEE}
\end{bchapter}
\endbibitem

\bibitem{long2015fully}
\begin{bchapter}
\bauthor{\bsnm{Long}, \binits{J.}},
\bauthor{\bsnm{Shelhamer}, \binits{E.}},
\bauthor{\bsnm{Darrell}, \binits{T.}}:
\bctitle{Fully convolutional networks for semantic segmentation}.
In: \bbtitle{Proceedings of the IEEE Conference on Computer Vision and Pattern Recognition},
pp. \bfpage{3431}--\blpage{3440}
(\byear{2015})
\end{bchapter}
\endbibitem

\bibitem{lee2017vpgnet}
\begin{bchapter}
\bauthor{\bsnm{Lee}, \binits{S.}},
\bauthor{\bsnm{Kim}, \binits{J.}},
\bauthor{\bsnm{Shin~Yoon}, \binits{J.}},
\bauthor{\bsnm{Shin}, \binits{S.}},
\bauthor{\bsnm{Bailo}, \binits{O.}},
\bauthor{\bsnm{Kim}, \binits{N.}},
\bauthor{\bsnm{Lee}, \binits{T.-H.}},
\bauthor{\bsnm{Seok~Hong}, \binits{H.}},
\bauthor{\bsnm{Han}, \binits{S.-H.}},
\bauthor{\bsnm{So~Kweon}, \binits{I.}}:
\bctitle{Vpgnet: Vanishing point guided network for lane and road marking detection and recognition}.
In: \bbtitle{Proceedings of the IEEE International Conference on Computer Vision},
pp. \bfpage{1947}--\blpage{1955}
(\byear{2017})
\end{bchapter}
\endbibitem

\bibitem{zou2019robust}
\begin{barticle}
\bauthor{\bsnm{Zou}, \binits{Q.}},
\bauthor{\bsnm{Jiang}, \binits{H.}},
\bauthor{\bsnm{Dai}, \binits{Q.}},
\bauthor{\bsnm{Yue}, \binits{Y.}},
\bauthor{\bsnm{Chen}, \binits{L.}},
\bauthor{\bsnm{Wang}, \binits{Q.}}:
\batitle{Robust lane detection from continuous driving scenes using deep neural networks}.
\bjtitle{IEEE transactions on vehicular technology}
\bvolume{69}(\bissue{1}),
\bfpage{41}--\blpage{54}
(\byear{2019})
\end{barticle}
\endbibitem

\bibitem{alvarez2014combining}
\begin{barticle}
\bauthor{\bsnm{Alvarez}, \binits{J.M.}},
\bauthor{\bsnm{L{\'o}pez}, \binits{A.M.}},
\bauthor{\bsnm{Gevers}, \binits{T.}},
\bauthor{\bsnm{Lumbreras}, \binits{F.}}:
\batitle{Combining priors, appearance, and context for road detection}.
\bjtitle{IEEE Transactions on Intelligent Transportation Systems}
\bvolume{15}(\bissue{3}),
\bfpage{1168}--\blpage{1178}
(\byear{2014})
\end{barticle}
\endbibitem

\bibitem{yang2021capturing}
\begin{bchapter}
\bauthor{\bsnm{Yang}, \binits{K.}},
\bauthor{\bsnm{Zhang}, \binits{J.}},
\bauthor{\bsnm{Rei{\ss}}, \binits{S.}},
\bauthor{\bsnm{Hu}, \binits{X.}},
\bauthor{\bsnm{Stiefelhagen}, \binits{R.}}:
\bctitle{Capturing omni-range context for omnidirectional segmentation}.
In: \bbtitle{Proceedings of the IEEE/CVF Conference on Computer Vision and Pattern Recognition},
pp. \bfpage{1376}--\blpage{1386}
(\byear{2021})
\end{bchapter}
\endbibitem

\bibitem{yang2019pass}
\begin{barticle}
\bauthor{\bsnm{Yang}, \binits{K.}},
\bauthor{\bsnm{Hu}, \binits{X.}},
\bauthor{\bsnm{Bergasa}, \binits{L.M.}},
\bauthor{\bsnm{Romera}, \binits{E.}},
\bauthor{\bsnm{Wang}, \binits{K.}}:
\batitle{Pass: Panoramic annular semantic segmentation}.
\bjtitle{IEEE Transactions on Intelligent Transportation Systems}
\bvolume{21}(\bissue{10}),
\bfpage{4171}--\blpage{4185}
(\byear{2019})
\end{barticle}
\endbibitem

\bibitem{yang2020omnisupervised}
\begin{botherref}
\oauthor{\bsnm{Yang}, \binits{K.}},
\oauthor{\bsnm{Hu}, \binits{X.}},
\oauthor{\bsnm{Fang}, \binits{Y.}},
\oauthor{\bsnm{Wang}, \binits{K.}},
\oauthor{\bsnm{Stiefelhagen}, \binits{R.}}:
Omnisupervised omnidirectional semantic segmentation.
IEEE Transactions on Intelligent Transportation Systems
(2020)
\end{botherref}
\endbibitem

\bibitem{xiao2018hybrid}
\begin{barticle}
\bauthor{\bsnm{Xiao}, \binits{L.}},
\bauthor{\bsnm{Wang}, \binits{R.}},
\bauthor{\bsnm{Dai}, \binits{B.}},
\bauthor{\bsnm{Fang}, \binits{Y.}},
\bauthor{\bsnm{Liu}, \binits{D.}},
\bauthor{\bsnm{Wu}, \binits{T.}}:
\batitle{Hybrid conditional random field based camera-lidar fusion for road detection}.
\bjtitle{Information Sciences}
\bvolume{432},
\bfpage{543}--\blpage{558}
(\byear{2018})
\end{barticle}
\endbibitem

\bibitem{caltagirone2017fast}
\begin{bchapter}
\bauthor{\bsnm{Caltagirone}, \binits{L.}},
\bauthor{\bsnm{Scheidegger}, \binits{S.}},
\bauthor{\bsnm{Svensson}, \binits{L.}},
\bauthor{\bsnm{Wahde}, \binits{M.}}:
\bctitle{Fast lidar-based road detection using fully convolutional neural networks}.
In: \bbtitle{2017 Ieee Intelligent Vehicles Symposium (iv)},
pp. \bfpage{1019}--\blpage{1024}
(\byear{2017}).
\bcomment{IEEE}
\end{bchapter}
\endbibitem

\bibitem{wang2017embedding}
\begin{barticle}
\bauthor{\bsnm{Wang}, \binits{Q.}},
\bauthor{\bsnm{Gao}, \binits{J.}},
\bauthor{\bsnm{Yuan}, \binits{Y.}}:
\batitle{Embedding structured contour and location prior in siamesed fully convolutional networks for road detection}.
\bjtitle{IEEE Transactions on Intelligent Transportation Systems}
\bvolume{19}(\bissue{1}),
\bfpage{230}--\blpage{241}
(\byear{2017})
\end{barticle}
\endbibitem

\bibitem{CFECA}
\begin{barticle}
\bauthor{\bsnm{Zhang}, \binits{X.}},
\bauthor{\bsnm{Li}, \binits{Z.}},
\bauthor{\bsnm{Gao}, \binits{X.}},
\bauthor{\bsnm{Jin}, \binits{D.}},
\bauthor{\bsnm{Li}, \binits{J.}}:
\batitle{Channel attention in lidar-camera fusion for lane line segmentation}.
\bjtitle{Pattern Recognition}
\bvolume{118},
\bfpage{108020}
(\byear{2021})
\end{barticle}
\endbibitem

\bibitem{SamalKSWM22}
\begin{barticle}
\bauthor{\bsnm{Samal}, \binits{K.}},
\bauthor{\bsnm{Kumawat}, \binits{H.}},
\bauthor{\bsnm{Saha}, \binits{P.}},
\bauthor{\bsnm{Wolf}, \binits{M.}},
\bauthor{\bsnm{Mukhopadhyay}, \binits{S.}}:
\batitle{Task-driven rgb-lidar fusion for object tracking in resource-efficient autonomous system}.
\bjtitle{{IEEE} Trans. Intell. Veh.}
\bvolume{7}(\bissue{1}),
\bfpage{102}--\blpage{112}
(\byear{2022}).
\doiurl{10.1109/TIV.2021.3087664}
\end{barticle}
\endbibitem

\bibitem{ZhouDLY23}
\begin{barticle}
\bauthor{\bsnm{Zhou}, \binits{W.}},
\bauthor{\bsnm{Dong}, \binits{S.}},
\bauthor{\bsnm{Lei}, \binits{J.}},
\bauthor{\bsnm{Yu}, \binits{L.}}:
\batitle{Mtanet: Multitask-aware network with hierarchical multimodal fusion for {RGB-T} urban scene understanding}.
\bjtitle{{IEEE} Trans. Intell. Veh.}
\bvolume{8}(\bissue{1}),
\bfpage{48}--\blpage{58}
(\byear{2023}).
\doiurl{10.1109/TIV.2022.3164899}
\end{barticle}
\endbibitem

\bibitem{ChenC22}
\begin{barticle}
\bauthor{\bsnm{Chen}, \binits{T.}},
\bauthor{\bsnm{Chang}, \binits{T.S.}}:
\batitle{Rangeseg: Range-aware real time segmentation of 3d lidar point clouds}.
\bjtitle{{IEEE} Trans. Intell. Veh.}
\bvolume{7}(\bissue{1}),
\bfpage{93}--\blpage{101}
(\byear{2022}).
\doiurl{10.1109/TIV.2021.3085827}
\end{barticle}
\endbibitem

\bibitem{SunLXS23}
\begin{barticle}
\bauthor{\bsnm{Sun}, \binits{Y.}},
\bauthor{\bsnm{Li}, \binits{J.}},
\bauthor{\bsnm{Xu}, \binits{X.}},
\bauthor{\bsnm{Shi}, \binits{Y.}}:
\batitle{Adaptive multi-lane detection based on robust instance segmentation for intelligent vehicles}.
\bjtitle{{IEEE} Trans. Intell. Veh.}
\bvolume{8}(\bissue{1}),
\bfpage{888}--\blpage{899}
(\byear{2023}).
\doiurl{10.1109/TIV.2022.3158750}
\end{barticle}
\endbibitem

\bibitem{fritsch2013new}
\begin{bchapter}
\bauthor{\bsnm{Fritsch}, \binits{J.}},
\bauthor{\bsnm{Kuehnl}, \binits{T.}},
\bauthor{\bsnm{Geiger}, \binits{A.}}:
\bctitle{A new performance measure and evaluation benchmark for road detection algorithms}.
In: \bbtitle{16th International IEEE Conference on Intelligent Transportation Systems (ITSC 2013)},
pp. \bfpage{1693}--\blpage{1700}
(\byear{2013}).
\bcomment{IEEE}
\end{bchapter}
\endbibitem

\bibitem{RESA}
\begin{botherref}
\oauthor{\bsnm{Zheng}, \binits{T.}},
\oauthor{\bsnm{Fang}, \binits{H.}},
\oauthor{\bsnm{Zhang}, \binits{Y.}},
\oauthor{\bsnm{Tang}, \binits{W.}},
\oauthor{\bsnm{Yang}, \binits{Z.}},
\oauthor{\bsnm{Liu}, \binits{H.}},
\oauthor{\bsnm{Cai}, \binits{D.}}:
Resa: Recurrent feature-shift aggregator for lane detection.
arXiv preprint arXiv:2008.13719
\textbf{5}(7)
(2020)
\end{botherref}
\endbibitem

\bibitem{hou2019learning}
\begin{bchapter}
\bauthor{\bsnm{Hou}, \binits{Y.}},
\bauthor{\bsnm{Ma}, \binits{Z.}},
\bauthor{\bsnm{Liu}, \binits{C.}},
\bauthor{\bsnm{Loy}, \binits{C.C.}}:
\bctitle{Learning lightweight lane detection cnns by self attention distillation}.
In: \bbtitle{Proceedings of the IEEE/CVF International Conference on Computer Vision},
pp. \bfpage{1013}--\blpage{1021}
(\byear{2019})
\end{bchapter}
\endbibitem

\bibitem{USNet}
\begin{bchapter}
\bauthor{\bsnm{Chang}, \binits{Y.}},
\bauthor{\bsnm{Xue}, \binits{F.}},
\bauthor{\bsnm{Sheng}, \binits{F.}},
\bauthor{\bsnm{Liang}, \binits{W.}},
\bauthor{\bsnm{Ming}, \binits{A.}}:
\bctitle{Fast road segmentation via uncertainty-aware symmetric network}.
In: \bbtitle{2022 International Conference on Robotics and Automation (ICRA)},
pp. \bfpage{11124}--\blpage{11130}
(\byear{2022}).
\bcomment{IEEE}
\end{bchapter}
\endbibitem

\bibitem{wang2021dynamic}
\begin{botherref}
\oauthor{\bsnm{Wang}, \binits{H.}},
\oauthor{\bsnm{Fan}, \binits{R.}},
\oauthor{\bsnm{Sun}, \binits{Y.}},
\oauthor{\bsnm{Liu}, \binits{M.}}:
Dynamic fusion module evolves drivable area and road anomaly detection: A benchmark and algorithms.
IEEE transactions on cybernetics
(2021)
\end{botherref}
\endbibitem

\bibitem{fan2021learning}
\begin{botherref}
\oauthor{\bsnm{Fan}, \binits{R.}},
\oauthor{\bsnm{Wang}, \binits{H.}},
\oauthor{\bsnm{Cai}, \binits{P.}},
\oauthor{\bsnm{Wu}, \binits{J.}},
\oauthor{\bsnm{Bocus}, \binits{M.J.}},
\oauthor{\bsnm{Qiao}, \binits{L.}},
\oauthor{\bsnm{Liu}, \binits{M.}}:
Learning collision-free space detection from stereo images: Homography matrix brings better data augmentation.
IEEE/ASME Transactions on Mechatronics,
225--233
(2021)
\end{botherref}
\endbibitem

\bibitem{gu2021cascaded}
\begin{bchapter}
\bauthor{\bsnm{Gu}, \binits{S.}},
\bauthor{\bsnm{Yang}, \binits{J.}},
\bauthor{\bsnm{Kong}, \binits{H.}}:
\bctitle{A cascaded lidar-camera fusion network for road detection}.
In: \bbtitle{2021 IEEE International Conference on Robotics and Automation (ICRA)},
pp. \bfpage{13308}--\blpage{13314}
(\byear{2021}).
\bcomment{IEEE}
\end{bchapter}
\endbibitem

\bibitem{LRDNet2022}
\begin{botherref}
\oauthor{\bsnm{Abdkhanstd}}:
LRDNet+.
GitHub
(2022)
\end{botherref}
\endbibitem

\bibitem{fan2020sne}
\begin{bchapter}
\bauthor{\bsnm{Fan}, \binits{R.}},
\bauthor{\bsnm{Wang}, \binits{H.}},
\bauthor{\bsnm{Cai}, \binits{P.}},
\bauthor{\bsnm{Liu}, \binits{M.}}:
\bctitle{Sne-roadseg: Incorporating surface normal information into semantic segmentation for accurate freespace detection}.
In: \bbtitle{European Conference on Computer Vision},
pp. \bfpage{340}--\blpage{356}
(\byear{2020}).
\bcomment{Springer}
\end{bchapter}
\endbibitem

\bibitem{wang2021sne}
\begin{bchapter}
\bauthor{\bsnm{Wang}, \binits{H.}},
\bauthor{\bsnm{Fan}, \binits{R.}},
\bauthor{\bsnm{Cai}, \binits{P.}},
\bauthor{\bsnm{Liu}, \binits{M.}}:
\bctitle{Sne-roadseg+: Rethinking depth-normal translation and deep supervision for freespace detection}.
In: \bbtitle{2021 IEEE/RSJ International Conference on Intelligent Robots and Systems (IROS)},
pp. \bfpage{1140}--\blpage{1145}
(\byear{2021}).
\bcomment{IEEE}
\end{bchapter}
\endbibitem

\end{thebibliography}

\end{document}